\documentclass[journal]{IEEEtran}
\ifCLASSINFOpdf
\else
\fi
\usepackage{dblfloatfix}
\usepackage{amssymb}
\usepackage{makecell}
\usepackage{subcaption}
\usepackage{booktabs}
\usepackage{upgreek}
\usepackage{verbatim} 
\usepackage{wrapfig}

\newcommand{\ie}{{\em i.e.}}
\newcommand{\eg}{{\em e.g.}}

\newcommand{\vs}{{\em vs}}
\newcommand{\Fig}[1]{Fig. \ref{fig:#1}}

\newcommand{\Sect}[1]{Sect. \ref{sec:#1}}

\usepackage{graphicx}
\graphicspath{{./figures/}}
\DeclareGraphicsExtensions{.jpg , .jpg, .eps, .png}

\usepackage[usernames]{xcolor}

\usepackage{hyperref}
\hypersetup{colorlinks=false, linkcolor=black, citecolor=black, urlcolor=cyan}

\begin{document}
%

\title{\LARGE \bf
Intention Recognition of Pedestrians and Cyclists by 2D Pose Estimation}
%
%
%



\author{Zhijie Fang, Antonio M. L\'opez
\thanks{Zhijie and Antonio are with the Dpt. of Computer Science at the Universitat Aut\`{o}noma de Barcelona (UAB), and with the Computer Vision Center (CVC) at the UAB. e-mail: \{zfang, antonio\}@cvc.uab.es}.
\thanks{We acknowledge the financial support by the Spanish project TIN2017-88709-R (MINECO/AEI/FEDER, UE). Antonio thanks the financial support by ICREA under the ICREA Academia Program. Zhijie thanks the Chinese Scholarship Council, Grant 201406150062. We thank the Generalitat de Catalunya CERCA Program and its ACCIO agency. We thank the student Arjun Gupta who annotated part of CASR during an internship at CVC.}
}

\maketitle

\begin{abstract}
Anticipating the intentions of vulnerable road users (VRUs) such as pedestrians and cyclists is critical for performing safe and comfortable driving maneuvers. This is the case for human driving and, thus, should be taken into account by systems providing any level of driving assistance, from advanced driver assistant systems (ADAS) to fully autonomous vehicles (AVs). In this paper, we show how the latest advances on monocular vision-based human pose estimation, {\ie} those relying on deep Convolutional Neural Networks (CNNs), enable to recognize the intentions of such VRUs. In the case of cyclists, we assume that they follow traffic rules to indicate future maneuvers with arm signals. In the case of pedestrians, no indications can be assumed. Instead, we hypothesize that the walking pattern of a pedestrian allows to determine if he/she has the intention of crossing the road in the path of the ego-vehicle, so that the ego-vehicle must maneuver accordingly ({\eg} slowing down or stopping). In this paper, we show how the same methodology can be used for recognizing pedestrians and cyclists' intentions. For pedestrians, we perform experiments on the JAAD dataset. For cyclists, we did not found an analogous dataset, thus, we created our own one by acquiring and annotating videos which we share with the research community. Overall, the proposed pipeline provides new state-of-the-art results on the intention recognition of VRUs.
\end{abstract}


%
\IEEEpeerreviewmaketitle

\section{INTRODUCTION}

\IEEEPARstart{E}{ven} there is still room to improve the detection and tracking of vulnerable road users (VRUs), {\ie} pedestrians and cyclists, the state-of-the-art is sufficiently mature \cite{Geronimo:2014, Ren:2015, Li:2016, Zhang:2017, Li:2017, Eldesokey:2017, Wojke:2017, Flohr:2018} as to allow for increasingly focus more on related higher level tasks, which are crucial in terms of assisted and automated driving safety and comfort. 

\begin{figure}
\centering
\includegraphics[width=0.7\columnwidth]{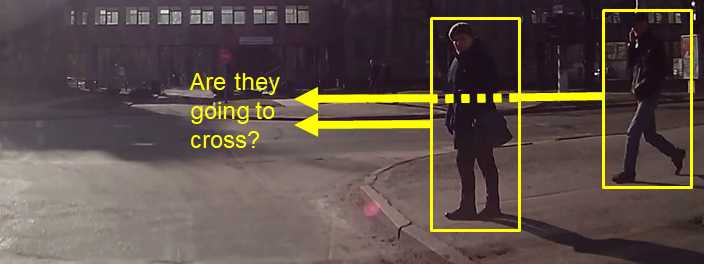}
\caption{Our focus: \emph{are the pedestrians going to cross?}}
\label{fig:cro-or-not-1frame}
\end{figure}

In particular, knowing the intention of a pedestrian to cross the road in front of the ego-vehicle, {\ie} before the pedestrian has actually entered the road, would allow the vehicle to warn the driver or automatically perform smoother maneuvers more respectful with pedestrians; it even significantly reduces the chance of injury requiring hospitalization when a vehicle-to-pedestrian crash is not fully avoidable \cite{Meinecke:2003}. The idea can be illustrated with the support of \Fig{cro-or-not-1frame}. We can see two pedestrians, one apparently stopped near a curb and the other walking towards the same curb. Just looking at the location of the (yellow) bounding boxes (BBs) that frame these pedestrians, we would say that they are not in the path of the vehicle at the moment. However, we would like to know what is going to happen next: is the stopped pedestrian suddenly going to cross the road? is the walking pedestrian going to cross the road without stopping?; in the affirmative cases, the vehicle could start to slow down already for a safer maneuver, increasing the comfort of the passengers and the confidence of the pedestrians (especially relevant for autonomous vehicles).

%

\begin{figure}
\centering
\includegraphics[clip=true, trim=300 000 150 060, width=0.188\columnwidth]{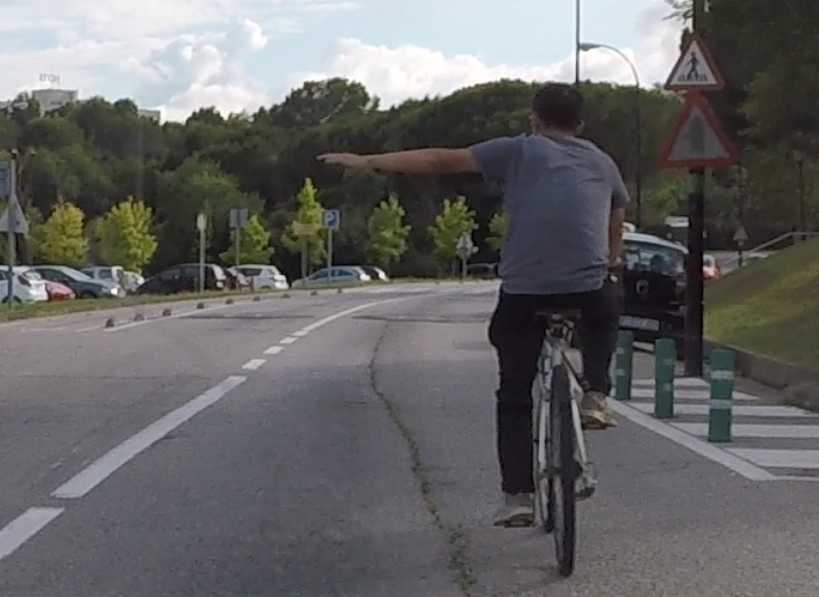}
\includegraphics[clip=true, trim=400 000 100 050, width=0.200\columnwidth]{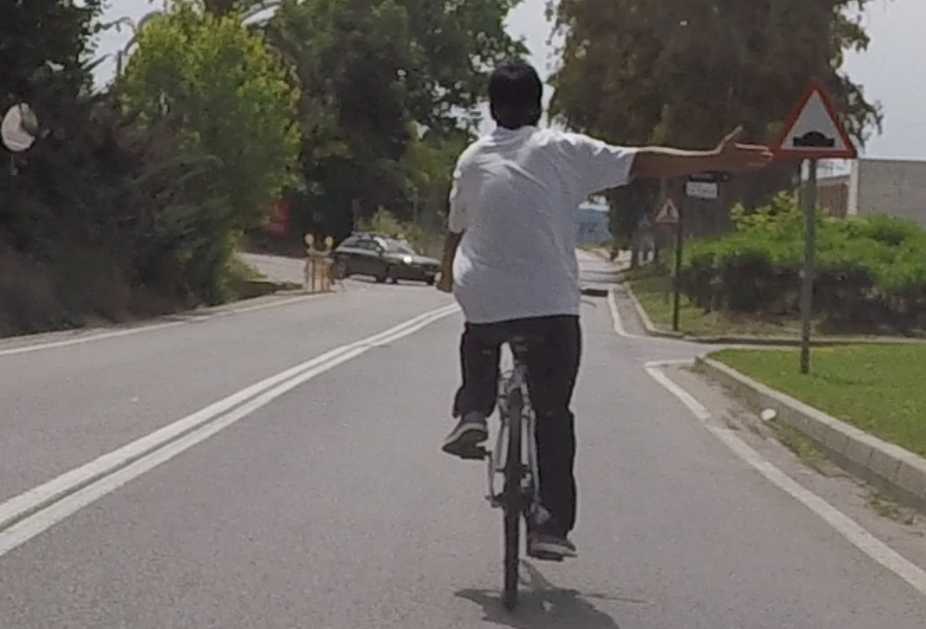}
\includegraphics[clip=true, trim=350 000 150 080, width=0.132\columnwidth]{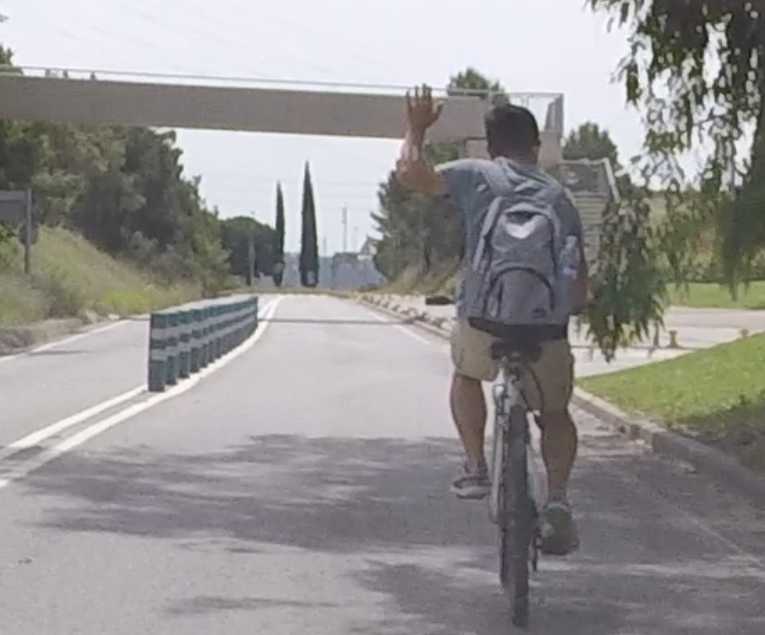}
\includegraphics[clip=true, trim=350 000 150 060, width=0.142\columnwidth]{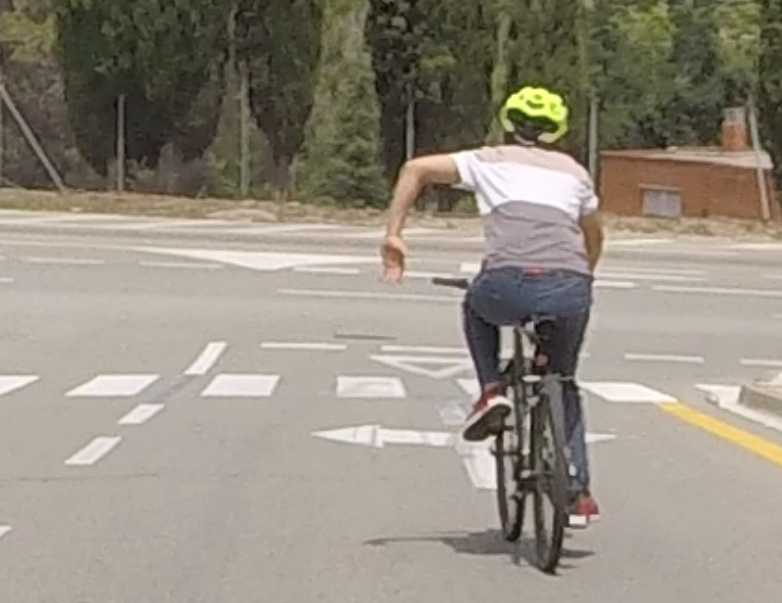}
\caption{Cyclist arm signals, left to right: Turn Left, Turn Right, Alternative Turn Right and Stop.}
\label{fig:arm-signals}
\end{figure}

Recognizing the motion intentions of cyclists is also highly relevant since many times the ego-vehicle will need to overtake them. While we cannot assume that pedestrians will explicitly indicate their intentions, for the cyclists we can exploit traffic rules. In particular, cyclists must indicate future left/right turns and stop maneuvers with arm signals (see \Fig{arm-signals}).    

In this paper, we explore the idea of using 2D pose estimation from monocular images as core information to recognize the intentions of both pedestrians and cyclists. In fact, we already addressed the pedestrian crossing/not-crossing classification (C/NC) task by relying on image-based 2D pose estimation \cite{Fang:2017, Fang:2018}. The proposed method shows state-of-the-art results and, in contrast to previous approaches (see \Sect{relatedWork}), it does not require information such as stereo, optical flow, or ego-motion compensation. In the most recent work \cite{Fang:2018}, we reported results for the (\emph{Joint Attention for Autonomous Driving}--JAAD) dataset \cite{Rasouli:2017}, which allows to address the C/NC classification task in naturalistic driving conditions. Moreover, since recently CNN-based features have been used to address the C/NC classification task in JAAD \cite{Rasouli:2017b}, we additionally compared our pose-estimation-based features with CNN-based ones, the former clearly outperforming the latter. 

Overall, \cite{Fang:2018} reported the new state-of-the-art baseline for JAAD. In this paper, first we complement this work by analysing the effect of obtaining noisy 2D skeletons, and we report also the features that the C/NC classifier considers most relevant. In addition, we show how the same methodology can be used to recognize cyclist arm signals; which also required to collect and annotate a specific dataset presented in this paper too. As in JAAD, our cyclist arm sign recognition (CASR) dataset was collected with a consumer-graded camera. We have annotated 219 arm signal actions on videos of approximately 10 seconds each, containing one or two actions per video. We also annotated 10 additional Youtube videos. CASR is publicly available in \url{https://github.com/VRU-intention/casr}. 

\Sect{relatedWork} reviews the related work. \Sect{method} presents our proposal to recognize VRU intentions. \Sect{experiments-pedestrians} and \Sect{experiments-cyclists} detail the experiments, results, and derived conclusions. We present CASR in \Sect{experiments-cyclists}. Finally, \Sect{conclusions} summarizes the work and its possible continuations. 
\section{RELATED WORK}
\label{sec:relatedWork}

C/NC classification started as a pedestrian path prediction problem; addressed by relying on pedestrian dynamic models for estimating pedestrian future location, speed and acceleration \cite{Schneider:2013, Keller:2014}. However, these models are difficult to adjust and for robustness require to rely on dense stereo data, dense optical flow and ego-motion compensation. Intuitively, methods like \cite{Keller:2014} implicitly try to predict how the silhouette of a tracked pedestrian evolves over time. In fact, \cite{Kohler:2015} uses a stereo-vision system and ego-motion compensation to explicitly assess the silhouette of the pedestrians (others rely on $360^\circ$ LIDAR \cite{Volz:2016}). Note that, while our method will be applied in JAAD because only relies on a monocular stream of images, these other methods cannot be applied due to the lack of stereo information and vehicle data for ego-motion compensation. 

On-board head and body orientation approximations have been also proposed to estimate pedestrian intentions, both from monocular \cite{Rehder:2014} and stereo \cite{Flohr:2015, Schulz:2015} images with ego-motion compensation. However, it is unclear how we actually can use these orientations to provide intention estimation. Moreover, the experiments reported in \cite{Schulz:2015} suggest that head detection is not useful for the C/NC classification task. 

These mentioned vision-based works relied on Daimler's dataset. By using an AlexNet-based CNN trained on JAAD, \cite{Rasouli:2017b} verified whether full body appearance improves the results on the C/NC classification task compared to analyzing only the sub-window containing either the head or the lower body. Conclusions were similar, {\ie} specifically focusing on legs or head does not seem to bring better performance. 

\begin{figure*}
\centering
\includegraphics[clip=true, trim=0 200 0 0, width=\textwidth]{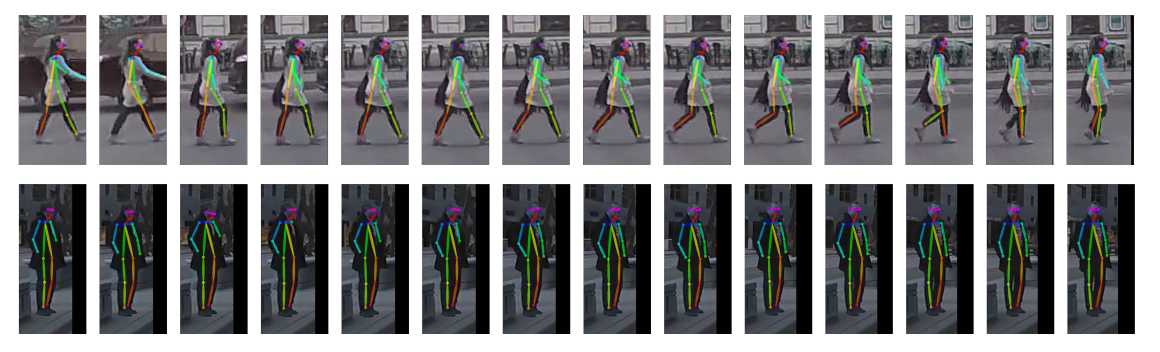}
\caption{2D skeleton fitting on 14 consecutive frames of one JAAD sequence, which roughly correspond to half a second.}
\label{fig:poseSequences}
\end{figure*}

In fact, \emph{a lack of information about the pedestrian's posture and body movement results in a delayed detection of the pedestrians changing their crossing intention} \cite{Schneemann:2016}. In line with this suggestion, in \cite{Fang:2017} we relied on a state-of-the-art 2D pose estimation method that operates in still images \cite{Cao:2017}. In particular, following a sliding time-window approach, accumulating estimated pedestrian skeletons over-time (see \Fig{poseSequences}) and features on top of these skeletons (see \Fig{skeleton_fets_display}), we obtained state-of-the-art results for the C/NC classification task in Daimler's dataset; which is remarkable since we only relied on a monocular stream of frames, neither on stereo, nor on optical flow, nor on ego-motion compensation. In this paper, we augment our study to the more challenging JAAD dataset by complementing the 2D pose estimation with state-of-the-art pedestrian detection and tracking. Moreover, we compare the use of skeleton-based features with CNN-appearance-based ones as suggested in \cite{Gkioxari:2014} for the generic task of human action recognition. We will see how the former bring more accuracy than the latter. We also report time-to-event results.

Pedestrian intention recognition is reviewed in \cite{Ridel:2018} for technologies such as LiDAR and wifi. Compared to pedestrian intention recognition, recognizing cyclist arm signals has received less attention. One reason may be the lack of publicly available datasets to assess this task. After \cite{Li:2016, Li:2017}, it was publicly released a large dataset focusing on cyclists, termed as Tsinghua-Daimler Cyclist Benchmark dataset (TDCB); however, acquired data and annotations are intended to support detection and orientation estimation tasks, but not cyclist arm signal recognition. In \cite{Eldesokey:2017}, the ground truth of TDCB was extended with wheel annotation for the case of bikes in side view, still to support cyclist detection. Thus, in this paper, we introduce our Cyclist Arm Signal Recognition dataset (CASR) containing 40,218 frames organized as short videos containing a cyclist arm signal each, in total, 219 annotated actions. For assessing generalization we also annotated 10 additional actions from Youtube, corresponding to 1,636 frames. 

 Using a stereo rig, in \cite{Flohr:2018, Kooij:2019} it is detected whether the left arm of a cyclist observed from the back is up or down, which is used as a context cue within a path prediction module. However, an isolated accuracy analysis of such up/down arm classification is not performed. In order to perform such a classification, the disparity map computed from stereo image pairs is used to produce a binary mask of each detected cyclist, and template matching is applied to determine if the mask correlates with a left arm up or down. In particular, the scores of matching against multiple templates, the disparity values, and the image intensities, are used as core information to build a Naive Bayesian Classifier with uniform prior, which is responsible for the desired up/down arm classification. In this paper, we do not assume stereo data and we not only account for the cyclist signal to turn left, but also to turn right (two types as shown in \Fig{arm-signals}) and stopping. Moreover, we apply exactly the same procedure for pedestrian intention recognition and for cyclist arm signal recognition. On the other hand, the classification output of our method could be also integrated as a cue for the path prediction module of \cite{Flohr:2018, Kooij:2019}. 
 
 Other concurrent works to ours, have been recently presented also exploring the use of skeleton-based features. In \cite{Quintero:2018}, Gaussian processes operating on fitted human skeletons are used to recognize four types of actions, namely walking, stopping, starting, and standing. Results on naturalistic publicly available driving datasets ({\ie} such as JAAD) are not reported and cyclist arm sign recognition is not addressed. In \cite{Ghori:2018}, fitted skeletons are analyzed along time using a long short-term memory (LSTM) for detecting actions such as  crossing, stopping, starting, turning, and walking along. In this case, sequences in the wild are used but, up to the best of our knowledge, they are not publicly available so it is difficult to compare them with JAAD for instance. On the other hand, cyclist arm sign recognition is not addressed. As an aside note, even we have not included in this paper, we also experimented with LSTMs but the obtained results were not better than the ones that we will report in this paper. In \cite{Deng:2018}, the main focus is to propose a human pose extraction method, the further analysis of how to use it for automatic action recognition is not considered, only a manually guided visual analysis is performed and, therefore, no quantitative results are reported.       
 
 Finally,  we would like to mention US Patent \cite{Kretzschmar:2015}. In the described approach, arm signal recognition is based on LiDAR data, while we rely only on monocular vision. \cite{Kretzschmar:2015} does not report results on any specific dataset; however, we think this LiDAR-based approach and ours can be complementary.

\section{METHOD}
\label{sec:method}

We need to detect pedestrians, track them, adjust a skeleton for each one (\Fig{poseSequences}), and apply a C/NC classifier that relies on skeleton-based features (\Fig{skeleton_fets_display}). Cyclist arm signal classification follows analogous steps. 

\paragraph{Detection} we can use a state-of-the-art pedestrian/cyclists (object) detector as long as it only requires a single RGB image as input, and returns a set of bounding boxes (BBs), each one framing a pedestrian/cyclist (by cyclist, we mean that the BB can frame only the rider or both the rider and the bike). Due to their popularity, here we have considered Faster R-CNN \cite{Ren:2015} and Mask R-CNN \cite{He:2017}, which can be found in both the Detectron \cite{Girshick:2018} and TensorPack \cite{wu2016tensorpack} frameworks.

\paragraph{Tracking} we selected \cite{Wojke:2017}, an efficient state-of-the-art multiple object tracker, with publicly available code. It uses the following \emph{state} for a detected object: $(u, v, \lambda, h, \dot{x}, \dot{y}, \dot{\lambda}, \dot{h})$; where $(u, v)$ is the central pixel of the BB, $\lambda$ its aspect ratio, $h$ its height, while $\dot{x}$, $\dot{y}$, $\dot{\lambda}$, and $\dot{h}$ are the respective derivatives over time. These variables are updated by Kalman filtering. For data association, it is used a cosine distance on top of CNN features (trained on a large-scale person re-identification dataset \cite{Zheng:2016}, thus, especially useful for tracking pedestrians and cyclists) which scores the degree of visual similarity between BB detections and predictions. A detection which does not have a high matching score with some prediction is pre-tracked; if the lack of matching holds during several consecutive frames, the track is consolidated as corresponding to a new detected object. Predictions which do not have a high matching score with a new detection during several frames are considered as disappeared objects (ended tracks). 

\paragraph{Skeleton fitting (pose estimation)} Given the good results obtained in \cite{Fang:2017}, we apply the CNN-based pose estimation method proposed in \cite{Cao:2017}, which has publicly available code. This method can operate in still monocular images and has been trained on the \emph{Microsoft COCO 2016 keypoints dataset} \cite{Lin:2014}. Thus, a priori it could be effective to fit the pose of both pedestrians and cyclists. In fact, this method is supposed to perform both human detection and pose estimation. However, as we will comment in \Sect{experiments-pedestrians}, for our problem it was more effective to rely on a detector-tracker pipeline and then run the pose estimation module only inside the tracked BBs, obtaining in that way the desired skeletons (see \Fig{poseSequences}).

\begin{figure}
\centering
\includegraphics[width=0.5\columnwidth]{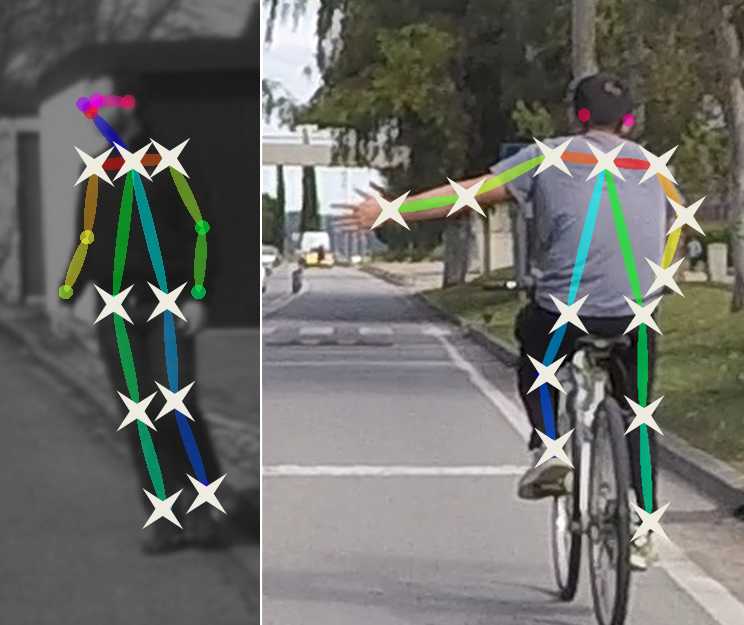}
\caption{Keypoints used for detecting the intentions of pedestrians (left) and cyclists (right). In the former case, 9 keypoints that are used out of the fitted skeleton to extract 396 features. In the later case, 13 keypoints are used to extract 1170 features.}
\label{fig:keypoints2}
\end{figure}

\paragraph{Intention classification} Focusing on the C/NC classification task, in \cite{Fang:2017} we extracted features from fitted pedestrian skeletons and use them as input to a shallow classifier. \Fig{keypoints2} highlights with stars the 9 keypoints we found as most stable, which correspond to the legs and the shoulders. These are highly relevant keypoints since the legs execute continue/start walking or stopping actions; while keypoints from shoulders and legs inform about global body orientation. From the selected keypoints we compute features. First, we perform a normalization of keypoint coordinates according to a factor $h$ proportional to the pedestrian \emph{height}, determined as the vertical distance from the top-most to the bottom-most keypoints (their location depends on the skeleton fitting). Then, different features (conveying redundant information) are computed by considering distances and relative angles between pairs of keypoints, as well as triangle angles induced by triplets of keypoints (see \Fig{skeleton_fets_display}). In total we obtain 396 features. Since we concatenate features during the last $T$ frames, our feature vector has dimension $396T$. Finally, we use a Random Forest (RF) classifier, which directly provides a probability for a meaningful thresholding to perform the C/NC classification. On the other hand, the pose of a cyclist is rather different than the pose of a pedestrian walking or standing. However, we hypothesize that, for performing arm signal classification, we can rely on the same keypoints than for C/NC classification plus the two additional keypoints from each arm (elbow and wrist, see \Fig{keypoints2}). Therefore, for each cyclist we use 13 keypoints, which turns out in 1170 features per frame, thus, $1170T$ for a temporal (sliding) window of $T$ frames.  

\section{PEDESTRIAN INTENTION EXPERIMENTS}
\label{sec:experiments-pedestrians}

\subsection{Dataset} 

First publicly available dataset for research on detecting pedestrian intentions is from Daimler \cite{Schneider:2013}. It contains 68 short sequences (9,135 frames in total) acquired in non naturalistic conditions and shows a single pedestrian per video, where the pedestrian performs pre-determined actions. More recently, it has been publicly released the Joint Attention for Autonomous Driving (JAAD) dataset \cite{Rasouli:2017}, acquired in naturalistic conditions and annotated for detecting C/NC actions. It contains 346 videos (most of them 5-10 seconds long) recorded on-board with a monocular system, running at 30 fps with a resolution of $1920\times1080$ pixels. Videos include both North America and Eastern Europe scenes. Overall, JAAD includes $\approx$~88,000 frames with 2,624 unique pedestrians labeled with $\approx$~390,000 BBs. Moreover, occlusion tags are provided for each BB. Where $\approx$~72,000 (18\%) BBs are tagged as partially occluded and $\approx$~46,000 (11\%) as heavily occluded. Pedestrian tracks are also provided. In addition, although we are not using it in this paper, JAAD contains also context information (traffic signs, street width, etc.) that we may use in further studies to complement purely pedestrian-based information. 

\subsection{Evaluation protocol}

In \cite{Rasouli:2017b}, JAAD was used for evaluating a C/NC classifier. However, neither is explained how JAAD was split in training and testing, nor the code is available. Here we use the first 250 videos of JAAD for training and the rest for testing. We have re-labeled as C the crossing labels of JAAD, as well as the labels in \{clear-path, moving-fast, moving-slow, slow-down, speed-up\} assigned to a pedestrian with lateral motion direction; the rest JAAD labels are re-labeled as NC.

\subsubsection{Training} 

While pedestrian detection based on the skeleton fitting algorithm that we use here \cite{Cao:2017} is possible, in our initial experiments we determined that its detection accuracy in JAAD was worse than using Faster R-CNN since the latter was fine-tuned from the available pedestrian BB annotations (fine-tuning the CNN of \cite{Cao:2017} would require body level annotations, not available in JAAD). In particular, we used the Faster R-CNN implementation released in \cite{Chen:2017}, following the same Faster R-CNN settings but using \{8, 16, 32, 64\} as anchors and 2.5 as BB aspect ratio ({\ie} pedestrian oriented). This Faster R-CNN is based on VGG16 CNN architecture. For fine-tuning we perform 110,000 iterations (remind that an iteration consists of a batch of 256 regions from the same image, and that input images are vertically mirrored to double the number of training samples). Regarding learning rate, we start with 0.001 and decrease the value to 0.0001 after 80,000 iterations.    

For training the C/NC classifier we rely on well-seen pedestrians and we balance the number of samples of the C and NC classes. Thus, we only consider pedestrian training samples with a minimum BB width of 60 pixels and no occlusion. Moreover, for a tracked pedestrian, these conditions must hold over more than T frames, since we need to concatenate last T frames for the C/NC classification. Thus, from tracks longer than T frames we can obtain different training samples by applying a temporal sliding window of size T. 

\begin{figure}[t!]
\centering
\includegraphics[width=\columnwidth]{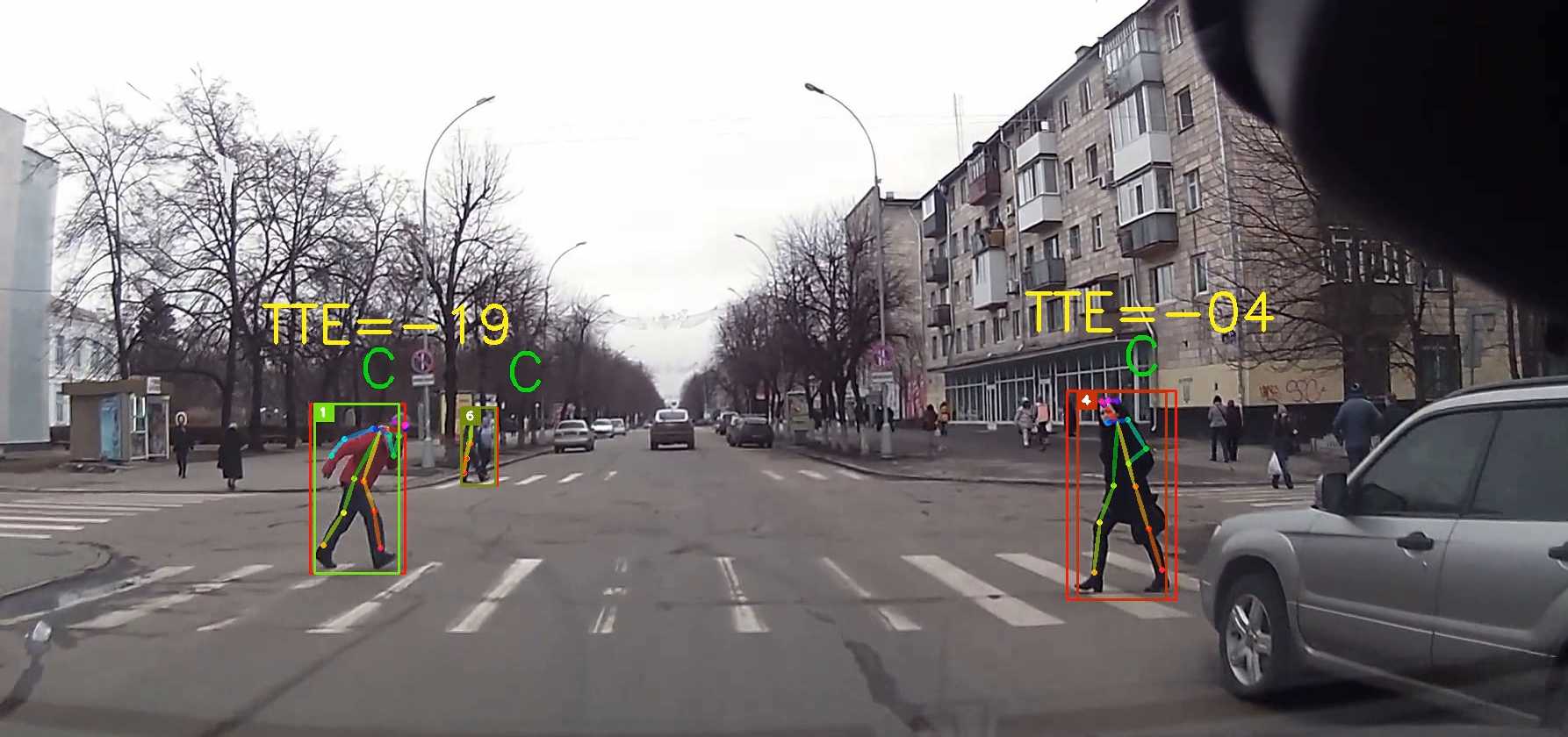}
\caption{C/NC classification. The ground truth label is indicated with "C" or "NC"; In green if the prediction agrees with the ground truth, red otherwise. Pedestrians are framed with two BBs: detection and tracking, the latter with the track ID. The estimated pedestrian skeleton is also shown. When annotated, time-to-event (TTE) is also shown in frame units. Negative TTE values mean that the event happened before this frame, while positive values indicate that it will happen after.}
\label{fig:res_c_nc_good}
\end{figure}

For each tracked pedestrian, the C/NC label assigned to a sequence of length T corresponds to the label in the most recent frame ({\ie} the frame in which the C/NC decision must be taken). We set T=14 for JAAD ({\ie}, following \cite{Fang:2017}, a value roughly below 0.5 seconds). Since we are in training time, here we are referring to the ground truth tracks provided in JAAD. For completeness, we also test the case T=1; meaning that we only train with the last frame of the same sequences used for the T=14 case. Overall, there are 8,677 sequences of length T=14 and NC label, while there are 36,253 with C label; thus, in the latter case we randomly take only 8,677 among those 36,253 possible. Accordingly, we fit the pose estimation-based skeleton and compute the C/NC features for 8,677 C and 8,677 NC samples (in a set of experiments for T=14, in another for T=1). These features are then used as input for the scikit-learn \cite{scikit-learn} function GridSearchCV; which is parameterized for training a Random Forest (RF) classifier using 5-fold cross-validation with the number of trees running on $\{100, 200, 300, 400, 500\}$ and maximum depth running on $\{7, 15, 21, 30\}$. The optimum RF in terms of accuracy corresponds to 400 trees and a maximum depth of 15, but we noted that all configurations provided similar accuracy.

Since we use Faster R-CNN, we can compare skeleton-based features with CNN-based ones, as is common in general action recognition literature \cite{Gkioxari:2014}. Accordingly, we apply the following procedure. For all training images we run the VGG16 obtained during Faster R-CNN fine-tuning. Then, for the same tracks mentioned before, we replace the skeleton-based features by the fc6 layer features inside the tracked pedestrian BBs. Note that (\Sect{method}) we have $396T$ skeleton-based features and $4096T$ fc6-based ones for each sample reaching RF training. In terms of RF parameter optimization (number of trees and maximum depth), CNN-based features reported similar accuracy as was the case for skeleton-based ones. Therefore, we set the same parameters, {\ie} 400 trees and a maximum depth of 15. We also combine skeleton and CNN-based features using the same RF parameters.

\subsubsection{Testing} 

In \cite{Rasouli:2017b}, evaluations are single frame (T=1) and only pedestrians with an action label are considered (mapped to C/NC here). When designing our experiments, we have seen that not all pedestrians of JAAD are annotated with a BB. Thus, when we run the detection and tracking modules, we are detecting and tracking some pedestrians which do not have the required ground truth information (BB, etc.). So, in order to follow a similar approach to \cite{Rasouli:2017b}, we do not consider these cases for quantitative evaluation. However, they are present in the qualitative evaluation ({\eg} see the videos provided as supplementary material). Overall, we ensure that T=1 and T=14 experiments are applied at the same tracked pedestrians at the same frames, performing a fair comparison. 

When detecting pedestrians with Faster R-CNN we use the default threshold $5\%$ and overlapping of $30\%$ for non-maximum suppression. For starting a new track, a pedestrian must be detected in 3 consecutive frames ({\ie} 0.1 seconds); while for ending a track there must be no new matched observations (detections) during 30 frames ({\ie} 1 second). For pose estimation (skeleton fitting) we use 3 scales; in particular, $\{1, (1-0.15), (1-0.15*2)\}$. For the C/NC classifier, we threshold in 0.5 the probability value provided by the RF.   

We assess accuracy according to the widespread definition $Acc = (TP+TN)/(P+N)$, where $P$ stands for total positives (here "C"), $N$ are the total negatives (here "NC"), and $TP$ and $TN$ the rightly classified positives and negatives (C and NC right classifications). According to the testing protocol we have defined, we found $P=17045$ and $N=5161$, therefore, $Acc$ could be bias towards "C" results. In order to avoid this, we select $P=N$ cases randomly. Thus, $Acc$ will be based on 10,322 testing decisions. 

As in \cite{Fang:2017}, we are interested in time-to-event (TTE) results for the critical case of crossing (C). However, JAAD is not annotated for this. Thus, we added TTE information to 9 \emph{keep-walking-to-cross} sequences, and to 14 \emph{start-walking-to-cross} ones. TTE = 0 is the when the event of interest happens. For \emph{keep-walking-to-cross}, it is the first frame at which the trunk of the walking pedestrian is over the curbside. For \emph{start-walking-to-cross}, it is the frame at which the stopped pedestrian starts moving a leg forward. 
Positive TTE values correspond to frames before the event, negative values to frames after the event. \Fig{res_c_nc_good} shows a result example where we can see TTE values for different pedestrians that are correctly classified as crossing (the supplementary videos have more examples). With TTE we provide two different plots, \emph{intention probability} {\vs} TTE, and \emph{predictability} {\vs} TTE. With the former we can see how many frames we can anticipate the pedestrian action. Since there are several testing sequences per intention, mean and standard deviation are plotted. \emph{Predictability} plots show a normalized measurement of how feasible is to detect the action under consideration for each TTE value. Predictability zero indicates that we cannot detect the action, while predictability one means that we can.  

\begin{table}
\hspace*{-0.2cm}
\caption{Classification accuracy (Acc) in JAAD. SKLT refers to our skeleton-based features, while CNN (fc6) are features from a VGG16 fine-tuned in JAAD (see main text). We have included here the results reported in \cite{Rasouli:2017b}, where CNN features are based on a non-fine-tuned AlexNet and Context refer to features of the environment, not of the pedestrian itself. Moreover, results for $20\%$ and $30\%$ noise in the keypoints is also reported for the SKLT case (se main text for details).}
\centering
\[
\begin{array}{cccccc}
\toprule
Method               &\textbf{T} & \textbf{features} & \textbf{Acc}  & \textbf{Acc} & \textbf{Acc} \\
                     &           &                   &               & 20\%         & 30\%         \\
\midrule
\cite{Rasouli:2017b} & 1         & CNN	             & 0.39          &              &                     \\
\cite{Rasouli:2017b} & 1         & CNN \& Context    & 0.63          &              &                     \\
\midrule
Ours                 & 1         & CNN (fc6)	     & 0.68          &              &                     \\
Ours                 & 1         & SKLT	             & 0.80          &  0.77        &       0.73          \\
Ours                 & 1         & CNN (fc6) + SKLT  & 0.81          &              &                     \\
\midrule
Ours                 & 14        & CNN (fc6)	     & 0.70          &              &                     \\
Ours                 & 14        & SKLT	             & \textbf{0.88} &  0.86        &       0.83          \\
Ours                 & 14        & CNN (fc6) + SKLT  & 0.87          &              &                     \\
\bottomrule
\end{array}
\]
\label{tab:accuracy}
\end{table}

\subsection{Results} 

Table \ref{tab:accuracy} reports accuracy results. We have included those reported in \cite{Rasouli:2017b}; however, our results are not directly comparable since it is unclear which frames were used for training and which ones for testing. The paper mentions that heavily occluded pedestrians are not considered for testing. In our experimental testing we do not exclude pedestrians due to occlusion. Moreover, we also report TTE information. However, we still found interesting to include the results in \cite{Rasouli:2017b} since the paper is based on CNN features and T=1. In particular, the authors train a walking/standing classifier and another looking/not-looking (pedestrian-to-car) classifier, both classifiers are based on a modified AlexNet CNN. The classification score of these classifiers is not used for final C/NC decision. Instead, the fc8 layer of both are used as features to perform a final C/NC based on a Linear-SVM adjusted in such a CNN-based feature space. It is also proposed to add contextual information captured by a place-classification style AlexNet.  

\begin{figure*}[t!]
\centering
\includegraphics[clip=true, trim = 125 000 195 000, width=0.245\textwidth]{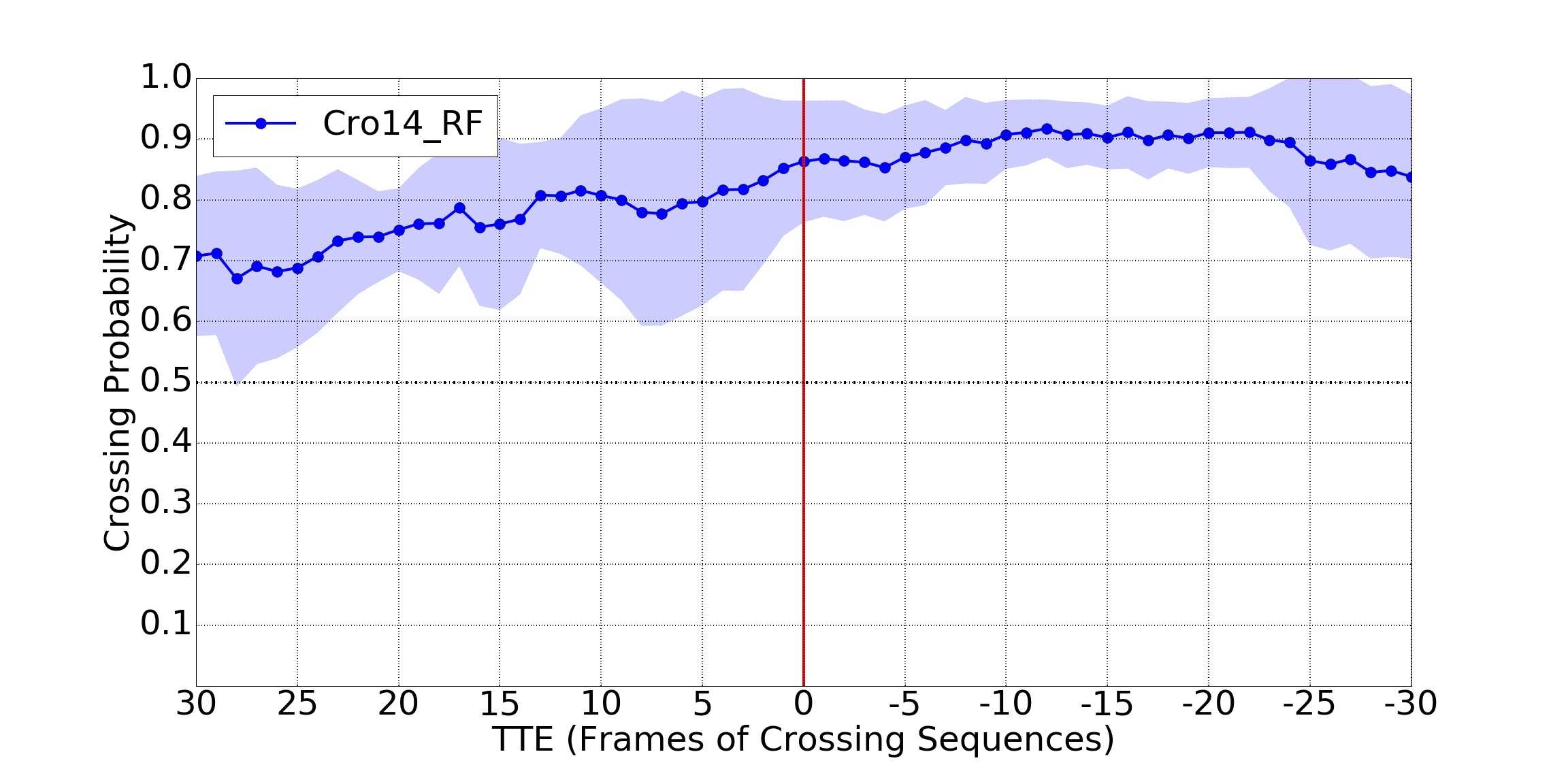}
\includegraphics[clip=true, trim = 125 000 195 000, width=0.245\textwidth]{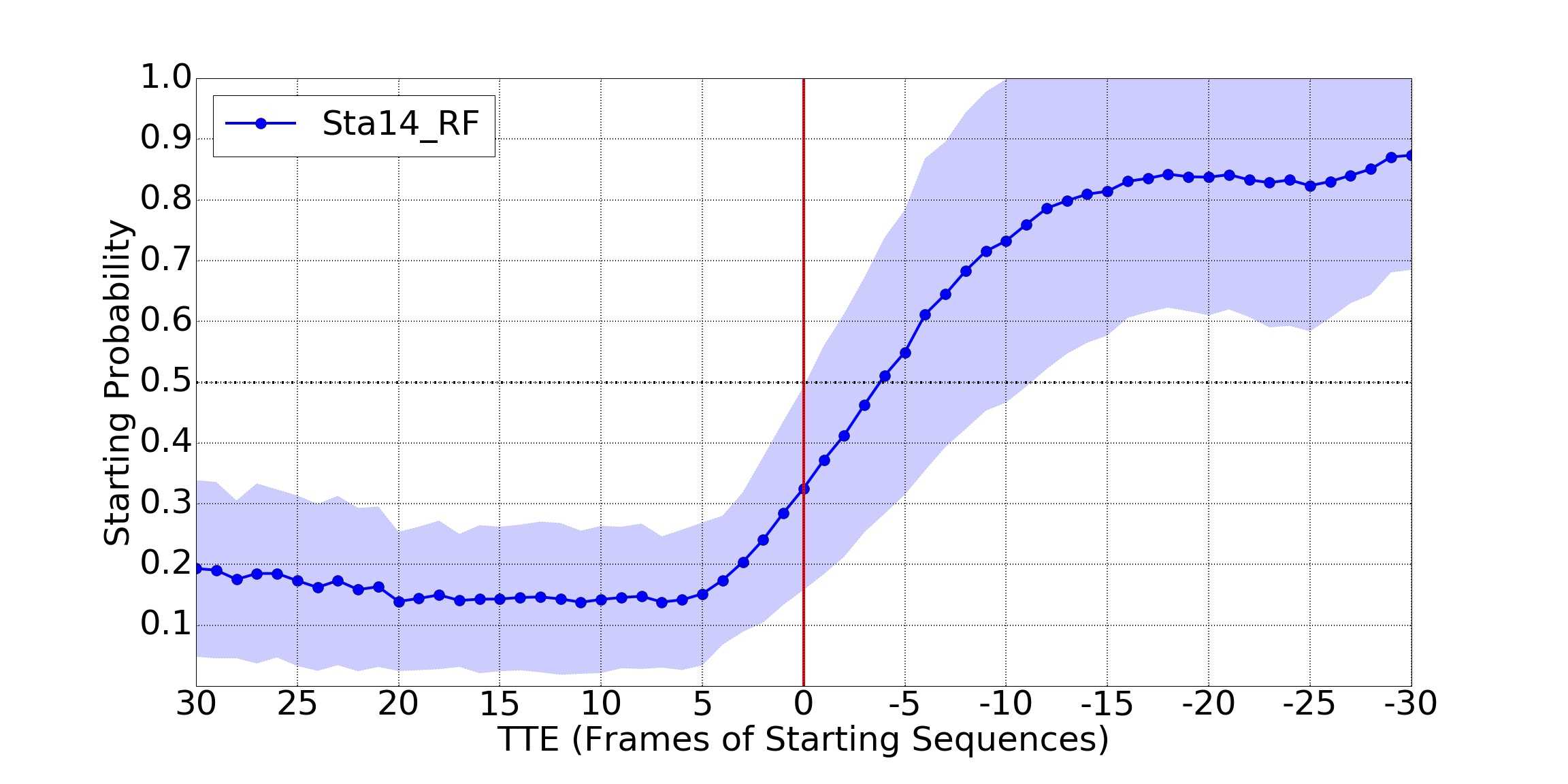}
\includegraphics[clip=true, trim = 125 000 195 000, width=0.245\textwidth]{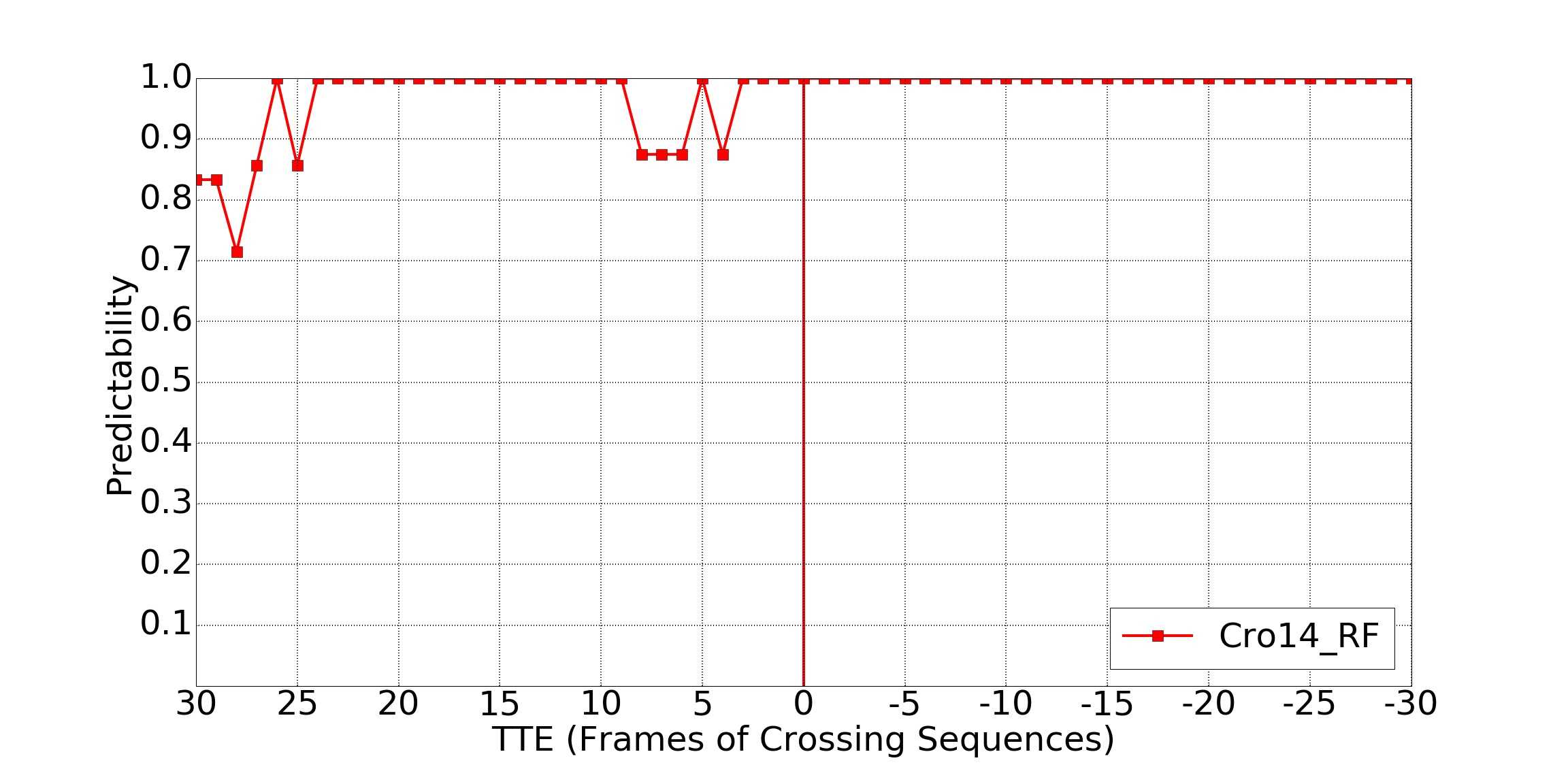}
\includegraphics[clip=true, trim = 125 000 195 000, width=0.245\textwidth]{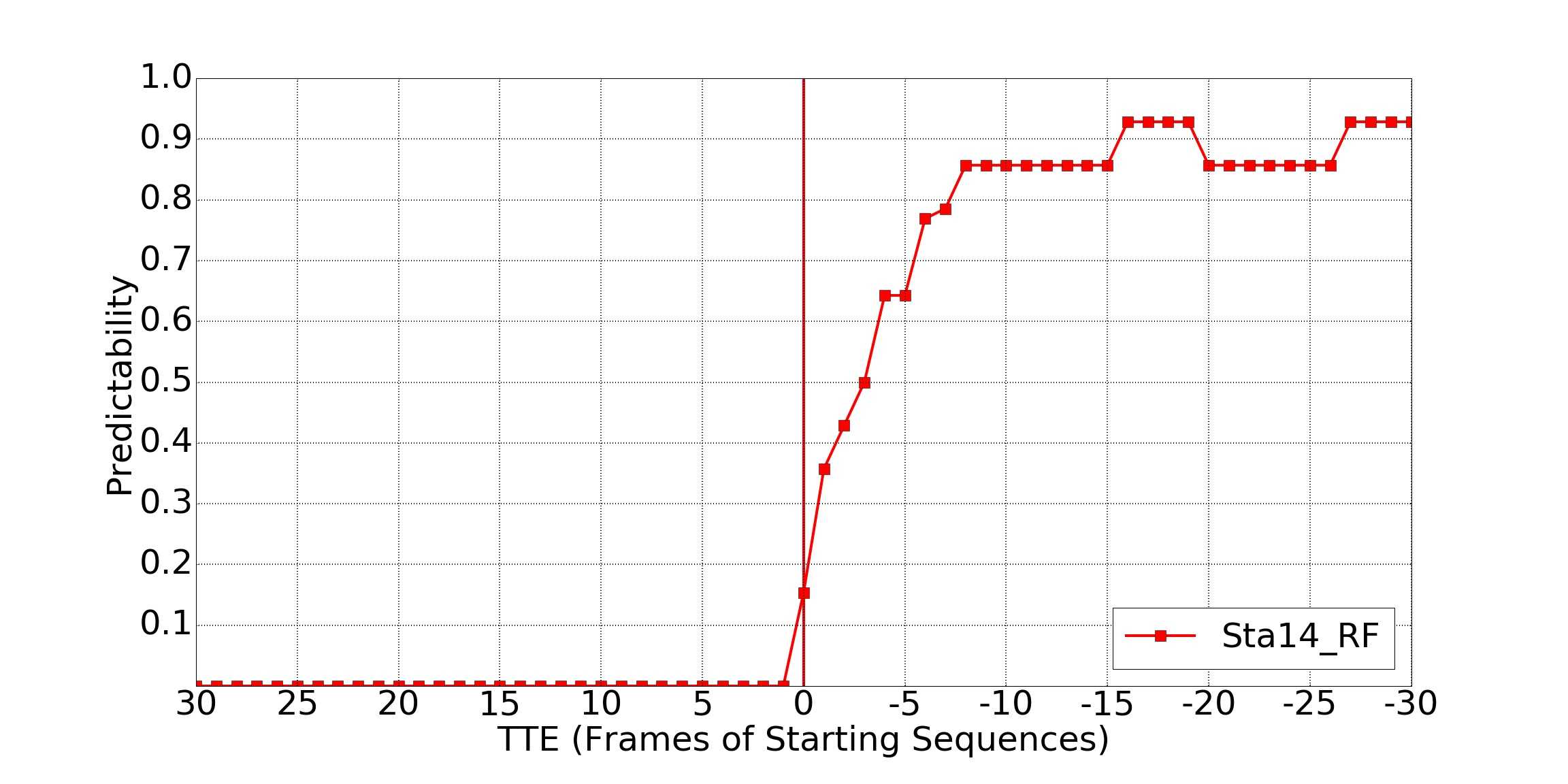}
\caption{For T=14, left to right: (a) Action prob. for \emph{Keep walking to cross}, mean over sequences and standard deviation; (b) Same for \emph{Start crossing}; (c) Predictability with action prob. thr. = 0.5, for \emph{Keep walking to cross}; (d) Same for \emph{Start crossing}.}
\label{fig:plotspredict}
\end{figure*}

For a fixed T, Table \ref{tab:accuracy} shows that the skeleton-based features (SKLT) outperform those based on CNN fc6 layer. Combining SKLT and fc6 does not significantly improves accuracy of SKLT. We can see also that T=14 outperforms T=1, showing the convenience of integrating different frames. \Fig{plotspredict} shows how the system is stable at predicting that a walking pedestrians will keep moving from a sidewalk and eventually crossing the curbside appearing in front of the vehicle. We can see also that we can predict (predictability$>$0.8) that a standing pedestrian will cross the curbside around 8 frames after he/she starts to move, which in JAAD is around 250ms. 

\begin{figure}[t!]
\centering
\includegraphics[clip=true, trim = 150 000 000 230, width=0.532\columnwidth]{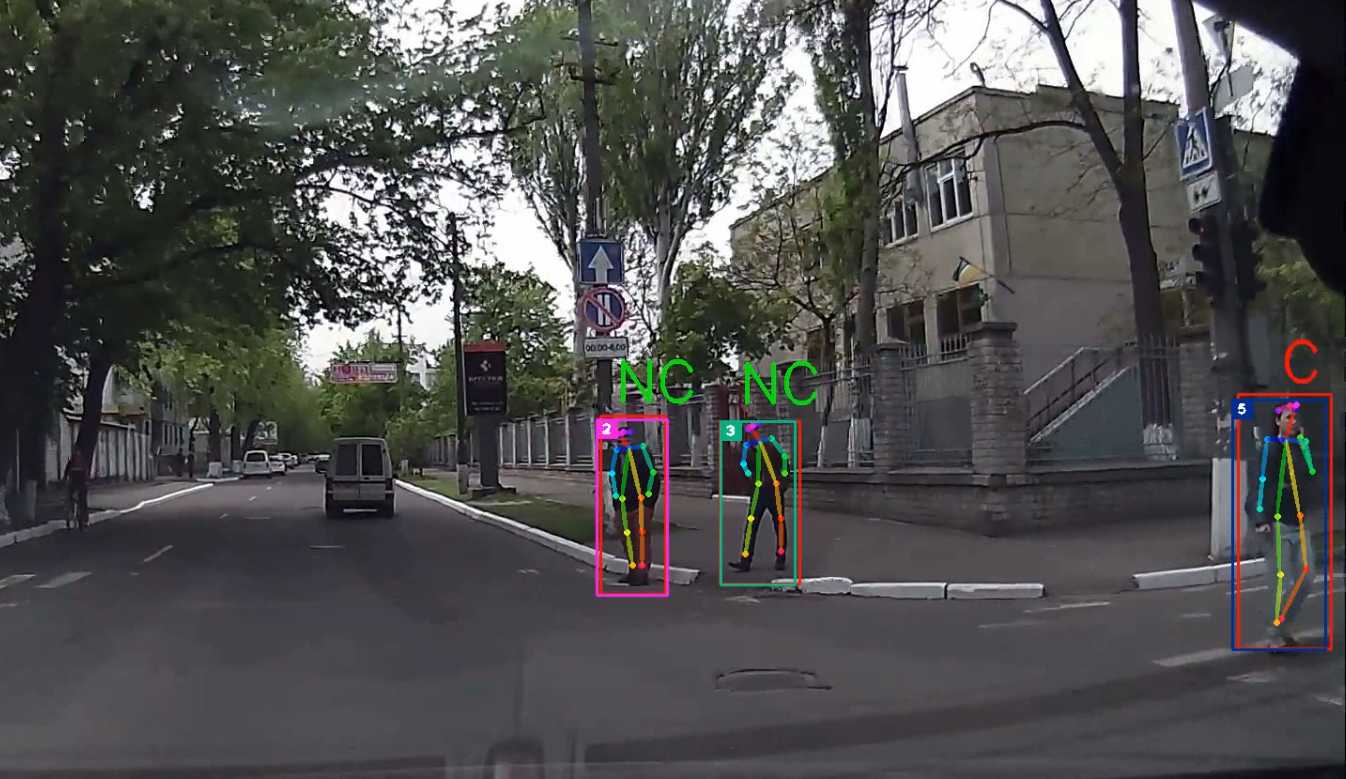}
\includegraphics[clip=true, trim = 050 000 300 150, width=0.452\columnwidth]{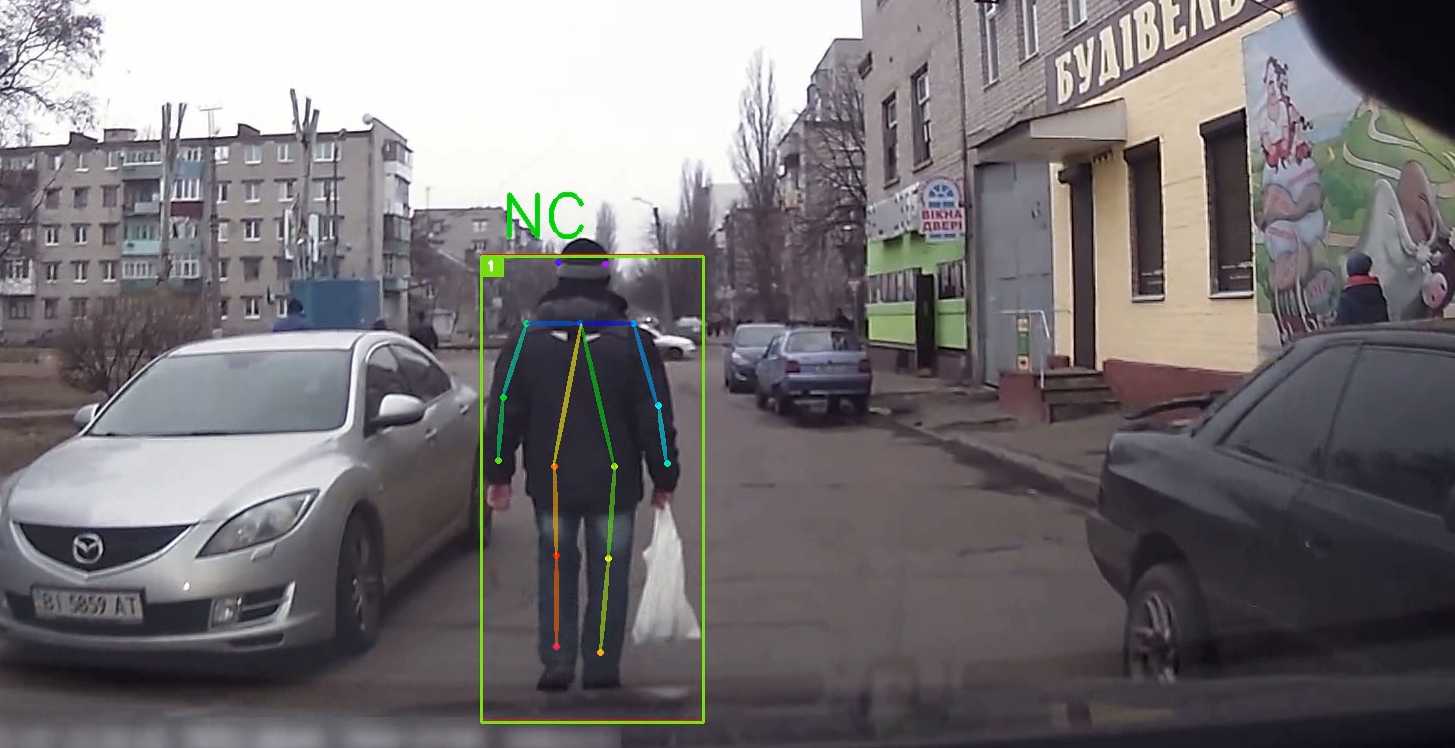}
\caption{Results of C/NC classification}
\label{fig:res_c_nc_2}
\end{figure}

Looking in more detail to the results, we find situations that need to be taken into account as future work. For instance, in \Fig{res_c_nc_2} there is a "C" accounted as error (red). Indeed, the pedestrian is crossing the road, but not the one intersecting the path of the ego-vehicle. So in the evaluation it should be probably accounted as right. On the contrary, in \Fig{res_c_nc_2} the system classifies as "NC" a pedestrian which is not crossing the road, but in fact is walking along the road, in front of the car. Now this situation is accounted as right, but probably should be accounted as wrong. On the other hand, in this case we can just use location-based reasoning to know that the pedestrian is in a dangerous place, it is not a problem of predicting the action anymore (as the C/NC case). It is worth also mentioning that we have observed that walking in parallel to the car motion direction, tends to be properly classified as NC; however, more annotations are required to provide a reasonable quantitative analysis. 

In order to evaluate the robustness of the method, we ran an equivalent set of experiments for the SKLT case, where we added random noise to the keypoints of the fitted skeleton in testing time. In particular,independently for each coordinate of each keypoint, we added Gaussian noise with zero mean and standard deviation $s$ which, following \cite{Jhuang:2013}, is set as a percentage over the distance to the closest keypoint. This is shown in Table \ref{tab:accuracy} for percentages of $20\%$ and $30\%$. As expected, accuracy decreases a few for $20\%$ and more for $30\%$, being T=14 is more robust to noise than T=1.

\begin{figure}[t!]
\centering
\includegraphics[width=0.6\columnwidth]{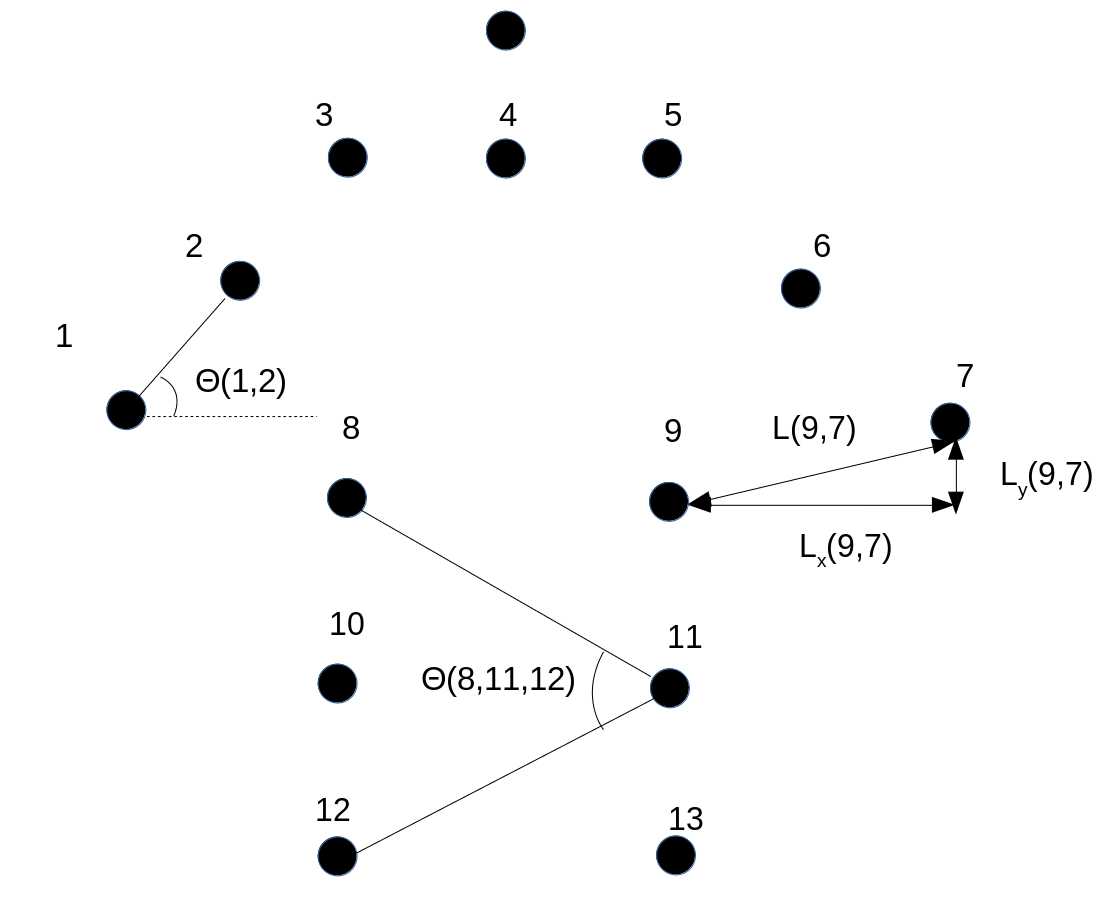}
\caption{Skeleton-based features: angles and distances.}
\label{fig:skeleton_fets_display}
\end{figure}

Finally, we assess the most important features for the RF classifier. In \Fig{skeleton_fets_display} we assign an ID to the keypoints of a fitted skeleton either used for pedestrian or cyclist intention recognition. The visualized naming scheme defines angles ($\Theta(\cdot,\cdot), \Theta(\cdot,\cdot,\cdot)$) and lengths ($\mbox{L}(\cdot,\cdot), \mbox{L}_x(\cdot,\cdot), \mbox{L}_y(\cdot,\cdot)$). Since we have evaluated both T=1 and T=14, in the latter case we also add a super-index to indicate from which relative frame index ({\ie} in $\{1,..,14\}$) comes the feature. Tables \ref{tab:relevancePedestrianFeaturesT1}-\ref{tab:relevancePedestrianFeaturesT14} show the top-25 more important features for T=1 (top $\sim 6\%$) and T=14 (top $\sim 0.5\%$), respectively. We see how all are based on 3-keypoint angles, mostly connecting either shoulder and legs, or shoulder and waist; thus, capturing global pose. For T=14, only one feature appears after frame 9 ($\sim 300\mbox{ms}$); thus, favoring intention prediction in a short time.

\begin{table}
\caption{For T=1, top-25 most relevant pedestrian skeleton-based features from left-to-right and top-to-bottom.}
\vspace{-0.3cm}
\centering
\begin{scriptsize}
\[
\begin{array}{ccccc}
\toprule
\Theta(4,12,3)  & \Theta(4,11,3)   & \Theta(8,4,9)   & \Theta(4,8,3)   & \Theta(3,10,5)   \\
\Theta(8,12,10) & \Theta(12,11,13) & \Theta(3,12,5)  & \Theta(4,13,3)  & \Theta(3,8,5)    \\
\Theta(4,10,3)  & \Theta(8,5,9)    & \Theta(8,3,9)   & \Theta(3,11,5)  & \Theta(12,10,13) \\
\Theta(4,9,3)   & \Theta(10,8,12)  & \Theta(3,9,5)   & \Theta(8,10,12) & \Theta(3,8,9)    \\
\Theta(3,9,8)   & \Theta(3,13,5)   & \Theta(9,13,11) & \Theta(12,9,13) & \Theta(12,8,13)  \\
\end{array}
\]
\end{scriptsize}
\label{tab:relevancePedestrianFeaturesT1}
\end{table}

\begin{table}
\caption{Analogous to Table \ref{tab:relevancePedestrianFeaturesT1} for T=14.}
\vspace{-0.3cm}
\centering
\begin{scriptsize}
\[
\begin{array}{ccccc}
\toprule
\Theta^7(4,9,3)  & \Theta^3(4,12,3)   & \Theta^8(4,13,3) & \Theta^9(4,12,3) & \Theta^2(4,12,3) \\
\Theta^6(4,9,3)  & \Theta^{12}(4,8,3) & \Theta^9(4,10,3) & \Theta^4(4,10,3) & \Theta^1(4,9,3)  \\
\Theta^4(4,8,3)  & \Theta^8(4,12,3)   & \Theta^6(4,8,3)  & \Theta^8(4,10,3) & \Theta^6(4,13,3) \\
\Theta^6(8,10,9) & \Theta^1(4,12,3)   & \Theta^4(3,8,5)  & \Theta^2(4,8,3)  & \Theta^3(4,8,3)  \\
\Theta^7(4,8,3)  & \Theta^4(4,9,3)    & \Theta^7(4,10,3) & \Theta^5(4,11,3) & \Theta^8(4,9,3)  \\
\end{array}
\]
\end{scriptsize}
\label{tab:relevancePedestrianFeaturesT14}
\end{table}

\section{CYCLIST INTENTION EXPERIMENTS}
\label{sec:experiments-cyclists}

\subsection{Dataset} 

\begin{table}[t!]
\caption{Cyclist arm signals in CASR and some YT videos.}
\centering
\begin{tabular}{|r|c|c|c|}
\hline
                & \textbf{Turn Left} & \textbf{Turn Right} & \textbf{Stop}\\
\hline
Cyclist  1		& 24	      & 38		    & 34 \\
Cyclist  2		& 16          & 24		    & 30 \\
Cyclist  3	    & 6	          & 12 	        & 26 \\
Cyclist  4		& 2	          & 3	        & 4	 \\
\hline
Total CASR      & \textbf{48} & \textbf{77} & \textbf{94} \\
\hline\hline
Total YouTube	& \textbf{6}  & \textbf{4}  & \textbf{0}  \\
\hline
\end{tabular}
\label{tab:sequences}
\end{table}

There are large datasets for cyclist detection such as the already mentioned TDCB \cite{Li:2016, Flohr:2018}. However, it does not include samples with annotations to assess arm signal recognition. Therefore, in this paper, we introduce our Cyclist Arm Signal Recognition dataset (CASR), consisting of 40,218 frames. Moreover, for assessing generalization we also annotated additional videos from YouTube, consisting of 1,626 frames. For CASR we followed a similar approach than JAAD authors. In particular, we attached a GoPro camera to the windshield of a car, forward facing the road ahead. We set the acquisition to RGB images at 30 fps and an $1920\times1080$ resolution. 

We asked four persons to drive their bikes inside our university campus, and they were instructed to ride around as they wish but using arm signals when required. Sometimes they wear helmet, sometimes not. Sometimes they carry a bag in their back, sometimes not. YouTube videos are also based on a dash cam facing the road. Table \ref{tab:sequences} summarizes the number of actions (cyclist arm signals) that we have annotated. Note that CASR includes 219 annotated actions, and YouTube 10. Actions have been organized as short videos of around 10 seconds with a single cyclist, where the frame starting an action and the frame ending this action are annotated. The videos of CASR mostly show one action and sometimes two actions because they were indicated in a continuous way by the cyclists, and in this case we did not split the video. In addition to the frame level action annotations, we have annotated the 2D BBs framing the cyclists too. Moreover, the videos where selected so that in most of them no pedestrians are included; thus, ready to focus on cyclist arm signal recognition. In some cases, however, there can be some pedestrians but we do not annotate his/her BB so that they are ignored during training and testing. Overall, CASR's content is analogous to the first deployed dataset for pedestrian intention recognition \cite{Schneider:2013}, but including much more annotated frames (68 actions within 9,135 in \cite{Schneider:2013}, 219 actions within 40,218 frames here).     

\begin{figure}
\centering
\begin{subfigure}{\columnwidth}
  \centering
  \makebox[\columnwidth][c]{
  \includegraphics[width=0.296\linewidth]{{hand-signal-right1}}
  \includegraphics[width=0.228\linewidth]{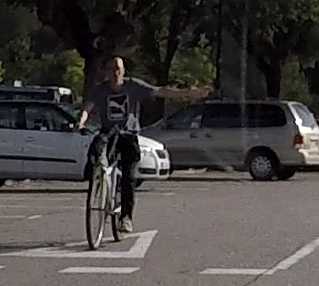}
  \includegraphics[width=0.252\linewidth]{{hand-signal-right2}}}
  \caption{Annotated as turning right}
  \label{fig:turning-right1-back-front}
\end{subfigure}
\begin{subfigure}{\columnwidth}
  \centering
  \makebox[\columnwidth][c]{
  \includegraphics[width=0.308\linewidth]{{hand-signal-left}}
  \includegraphics[width=0.244\linewidth]{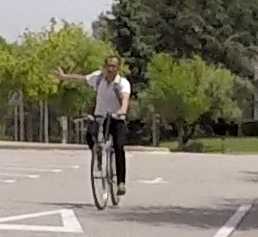}
  \includegraphics[width=0.228\linewidth]{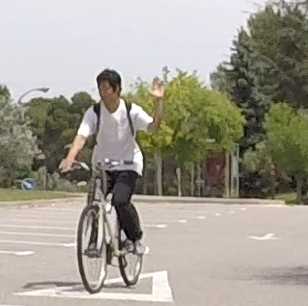}}
  \caption{Annotated as turning left}
  \label{fig:turning-left-back-front}
\end{subfigure}
\begin{subfigure}{\columnwidth}
  \centering
  \makebox[\columnwidth][c]{
  \includegraphics[width=0.292\linewidth]{{hand-signal-stop}}
  \includegraphics[width=0.256\linewidth]{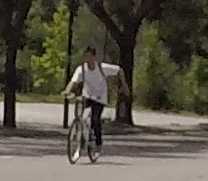}}
  \caption{Annotated as stopping}
  \label{fig:stop-back-front}
\end{subfigure}
\caption{Annotation of cyclist arm signals. We have followed a vehicle-centric criterion for left/right annotation.}
\label{fig:arm-signals-back-front}
\end{figure}

Note how action annotations are vehicle-centric here, instead of cyclist-centric. When the ego-vehicle follows the cyclist, they are the same. However, when the cyclist and the ego-vehicle move in opposite directions, we annotated as left-turn what for the cyclist is an indication of right-turn, and vice versa. The reason, is that for the vehicle what matters is the direction that the cyclist is going to take as seen in the image to be processed. Figure \ref{fig:arm-signals-back-front} clarifies the idea.    

\subsection{Evaluation protocol}
In CASR we recorded four cyclists. Accordingly, in order to evaluate our arm signal classifier, we divide their videos in training, validation, and testing sets. We use the videos of two cyclists for training, the videos of the other two cyclists are used for validation (training time) and testing, respectively. By varying the role of the cyclists, we can perform 12 training-validation-testing runs. Moreover, for each trained classifier, we test on the annotated YouTube videos too. We report individual metrics for each trained classifier, as well as averaged metrics. Since we aim at performing per-frame arm signal classification, we use the F1 and Accuracy standard metrics by counting classification errors and successes in each tested frame. For each training-validation run, we performed RF hyper-parameter search for the number of trees and the maximum depth allowed before performing the corresponding testing. For the former, we validated over the set $\{50, 100, 200, 300, 400, 500\}$, and for the later over the set $\{7, 10, 13, 16, 19, 22, 25, 28, 31, 34\}$. In this case, we also consider T=1 and T=14 as temporal sliding window sizes.

\begin{table}[t!]
\hspace{-0.2cm}
\caption{Classification accuracy (\textbf{Acc}) and \textbf{F1} score, both ranging from $0$ to $1$. Worst, best and average (with standard deviation)  results over 12 experimental runs for each T value. We  report generalization results on YouTube videos (\textbf{-YT}).}
\centering
\[
\begin{array}{ccccc}
\toprule
\mbox{T=1}   & \textbf{Acc} & \textbf{F1} & \textbf{Acc-YT} & \textbf{F1-YT}\\
\midrule
\mbox{Worst} & 0.89         & 0.86        & 0.82            & 0.78   \\
\mbox{Best}  & 0.96         & 0.96        & 0.83            & 0.77   \\
\mbox{Avg}   & 0.93\pm0.02  & 0.92\pm0.03 & 0.82\pm0.01     & 0.76\pm0.02 \\
\midrule
\mbox{T=14}  & \textbf{Acc} & \textbf{F1} & \textbf{Acc-YT} & \textbf{F1-YT} \\
\midrule
\mbox{Worst} & 0.89         & 0.87        & 0.83            & 0.79   \\
\mbox{Best}  & 0.96         & 0.96        & 0.85            & 0.79   \\
\mbox{Avg}   & 0.93\pm0.02  & 0.92\pm0.03 & 0.83\pm0.02     & 0.77\pm0.02  \\
\bottomrule
\end{array}
\]
\label{tab:accuracy_T1T14less}
\end{table}

\begin{table}[t!]
\hspace{-0.2cm}
\caption{Average classification accuracy (\textbf{Acc}) and \textbf{F1} score, for both CASR and YouTube videos (\textbf{-YT}). Results are reported for noise free keypoints, {\ie} using them as provided by the skeleton fitting algorithm, as well as for two different levels of noise ($20\%$ and $30\%$) on their location, which is forced at testing time (main text for details).}
\centering
\[
\begin{array}{ccccc}
\toprule
\mbox{T=1}  & \textbf{Acc} & \textbf{F1} & \textbf{Acc-YT} & \textbf{F1-YT} \\
\midrule
Noise free	& 0.93\pm0.02 & 0.92\pm0.03  & 0.82\pm0.01     & 0.76\pm0.02    \\
20\%		& 0.91\pm0.02 & 0.90\pm0.03  & 0.81\pm0.01     & 0.75\pm0.02    \\
30\%	    & 0.87\pm0.02 & 0.86\pm0.03  & 0.80\pm0.01     & 0.72\pm0.02    \\
\midrule
\mbox{T=14} & \textbf{Acc} & \textbf{F1} & \textbf{Acc-YT} & \textbf{F1-YT} \\
\midrule
Noise free	& 0.93\pm0.02 & 0.92\pm0.03  & 0.83\pm0.02     & 0.77\pm0.02    \\
20\%		& 0.93\pm0.02 & 0.92\pm0.02  & 0.83\pm0.02     & 0.77\pm0.02    \\
30\%	    & 0.92\pm0.02 & 0.92\pm0.02  & 0.83\pm0.02     & 0.76\pm0.02    \\
\bottomrule
\end{array}
\]
\label{tab:keypointNoiseResults}
\end{table}

Since CASR and the annotated YouTube videos contain a single cyclist and no pedestrians, detecting the cyclist is sufficient to perform our evaluation, {\ie} we do not need to run a tracker. Moreover, since the human pose estimation method that we use \cite{Cao:2017} performs the double task of searching the human and fitting its skeleton in a given 2D BB, we first relied on the fine-tuned Faster R-CNN detector described in \Sect{experiments-pedestrians} for providing such BBs. However, it tends to leave the arms of the cyclists out of the BBs. Thus, since our focus is not on detection,  we changed to Mask R-CNN \cite{He:2017}, in this case we did not fine-tuned the detector to CASR since object-level silhouette ground truth would be required. In practice, Mask R-CNN is providing accurate detections in CASR, and only in a few frames the corresponding detections were missing. In these frames, since we are not running a tracker, we just took a noisy version of the ground truth BB as detection. In particular, we added uniform noise to the BB's corner coordinates, being the amount of noise proportional to the size of the BB (we added independent Gaussian noise to each corner coordinates, with zero mean and st. dev. set to $10\%$ of the BB height, width, $x$ and $y$ coordinates). With this protocol we focus on the cyclist arm signal recognition itself. Moreover, \Sect{experiments-pedestrians} shows that the conclusions on pedestrian intention recognition that we draw in \cite{Fang:2017} using a dataset designed just for such a task, also hold in naturalistic traffic conditions. Thus, we expect the same for cyclist arm signal recognition. 

\begin{figure*}
\centering
  \includegraphics[width=0.112\textwidth]{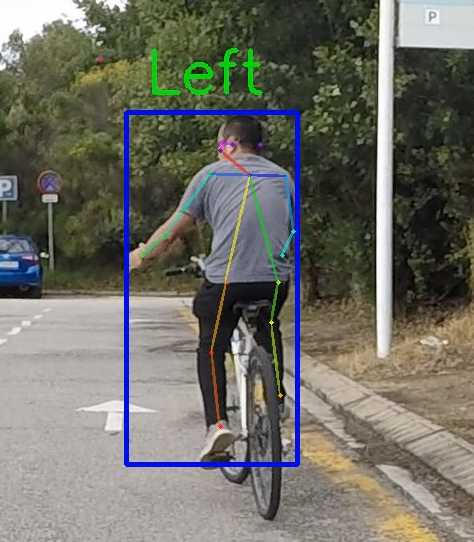}
  \includegraphics[width=0.140\textwidth]{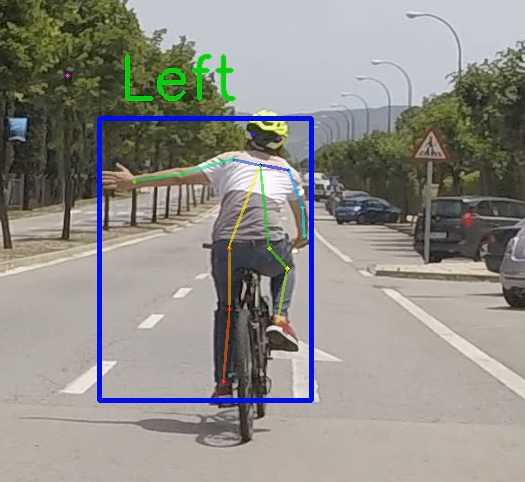}
  \includegraphics[width=0.123\textwidth]{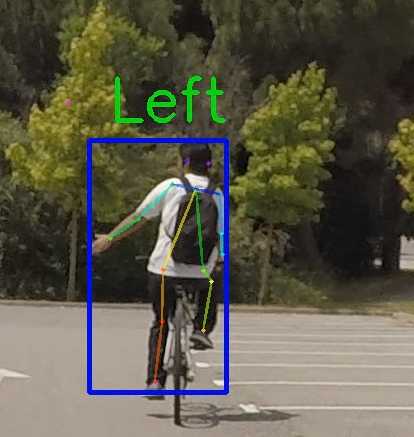}
  \includegraphics[width=0.138\textwidth]{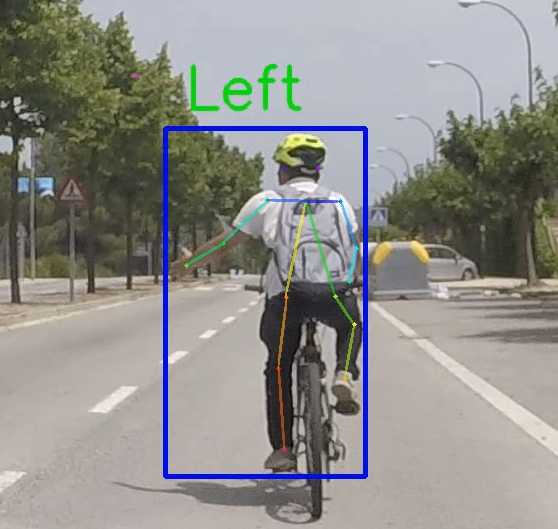}
  \includegraphics[width=0.128\textwidth]{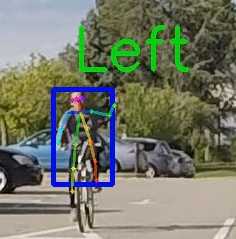}
  \includegraphics[width=0.135\textwidth]{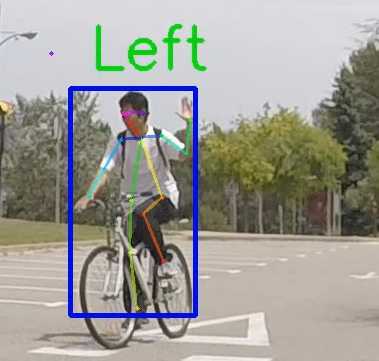}
\caption{Correct predictions in CASR for cyclist left turn indications (cropped from the original images). Remind that we are using a vehicle-centric criteria, this is why for oncoming cyclist an indication as right-turn must be classified as left-turn.}
\label{fig:leftOK}
\end{figure*}

\begin{figure*}
\centering
  \includegraphics[width=0.127\textwidth]{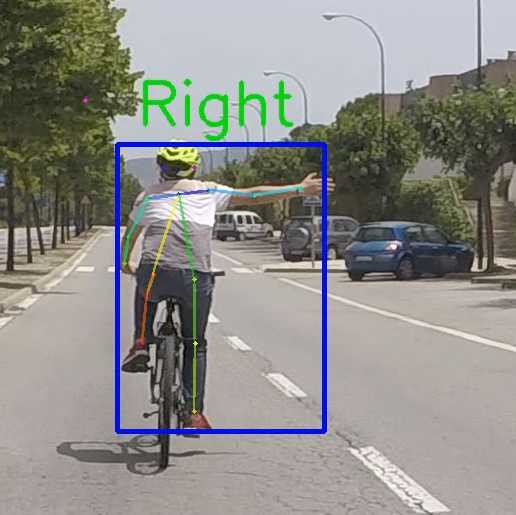}
  \includegraphics[width=0.118\textwidth]{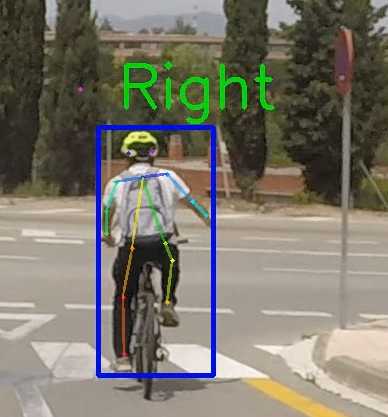}
  \includegraphics[width=0.122\textwidth]{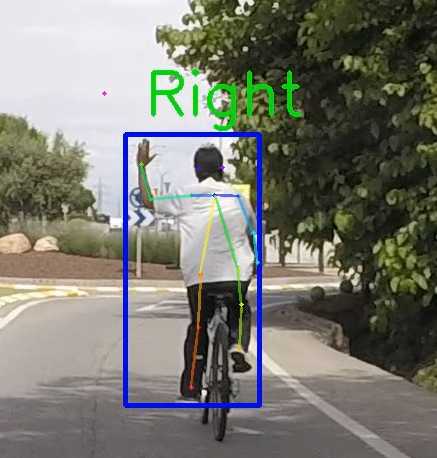}
  \includegraphics[width=0.138\textwidth]{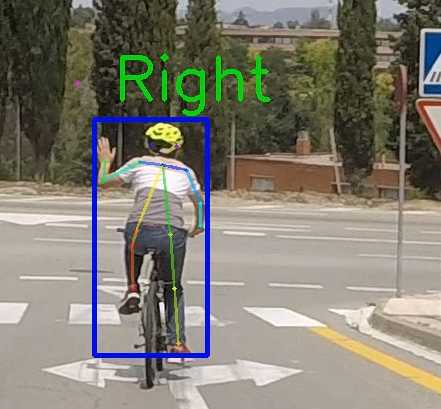}
  \includegraphics[width=0.138\textwidth]{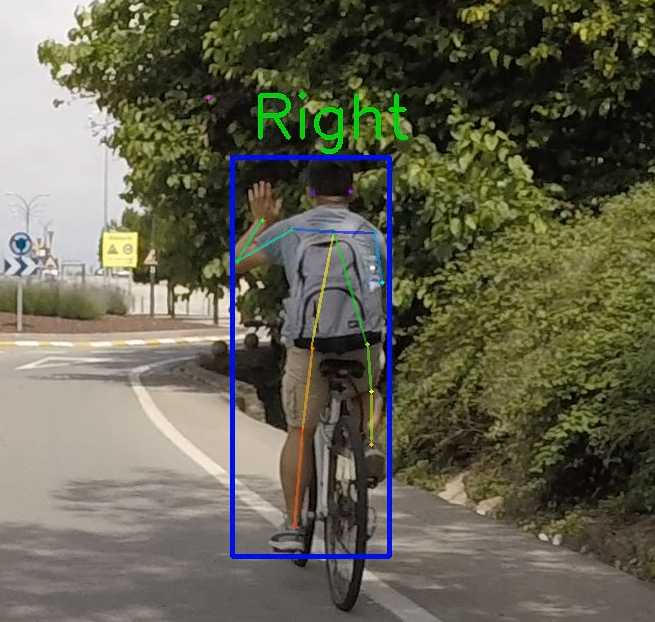}
  \includegraphics[width=0.126\textwidth]{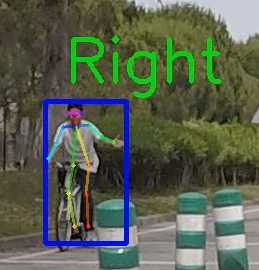}  
\caption{Correct predictions in CASR for cyclist right turn indications (cropped from the original images). Remind that we are using a vehicle-centric criteria, this is why for oncoming cyclist an indication as left-turn must be classified as right-turn.}
\label{fig:rightOK}
\end{figure*}

\begin{figure*}
 \centering
  \includegraphics[width=0.155\textwidth]{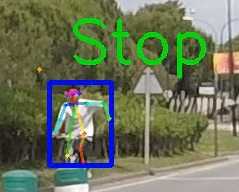}
  \includegraphics[width=0.127\textwidth]{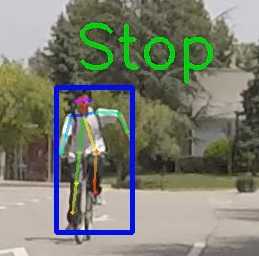}
  \includegraphics[width=0.127\textwidth]{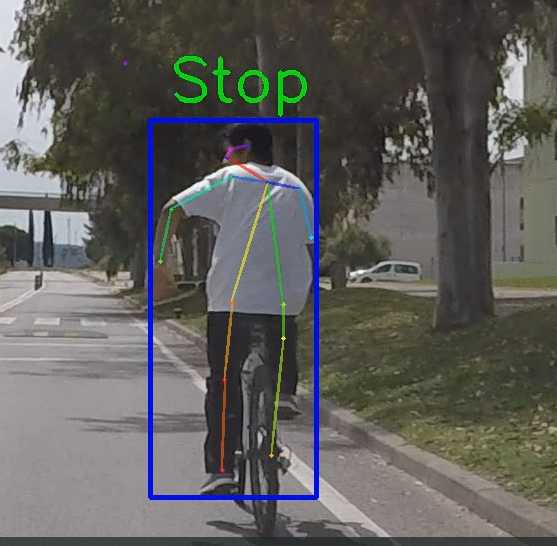}
  \includegraphics[width=0.127\textwidth]{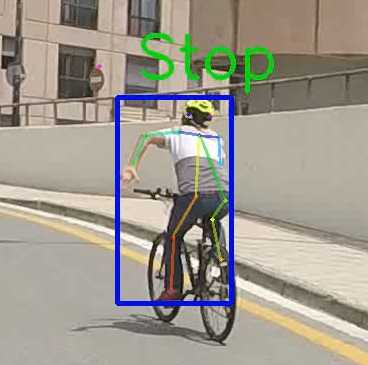}
  \includegraphics[width=0.127\textwidth]{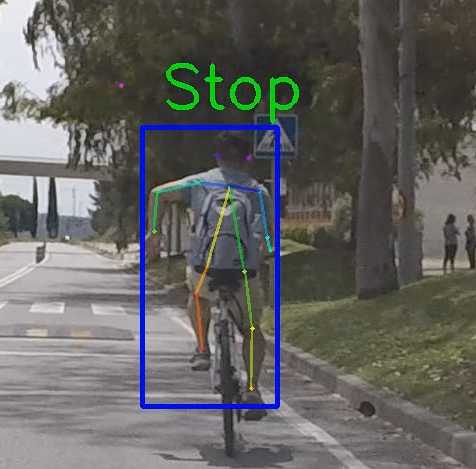}
  \includegraphics[width=0.115\textwidth]{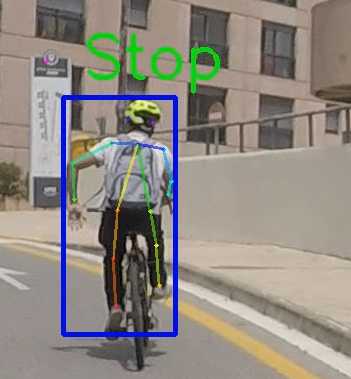}
\caption{Examples of correct predictions in CASR for cyclist stop indications (cropped from the original images).}
\label{fig:stopOK}
\end{figure*}

\begin{figure*}
\centering
  \includegraphics[width=0.152\textwidth]{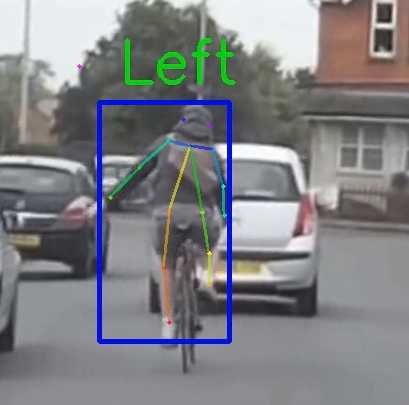}
  \includegraphics[width=0.164\textwidth]{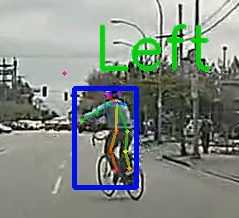}
  \includegraphics[width=0.160\textwidth]{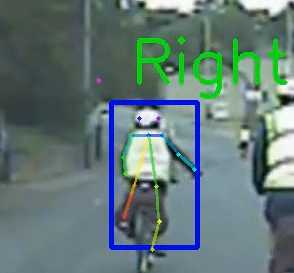}
  \includegraphics[width=0.148\textwidth]{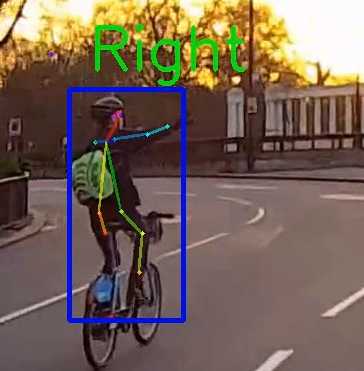}
  \includegraphics[width=0.136\textwidth]{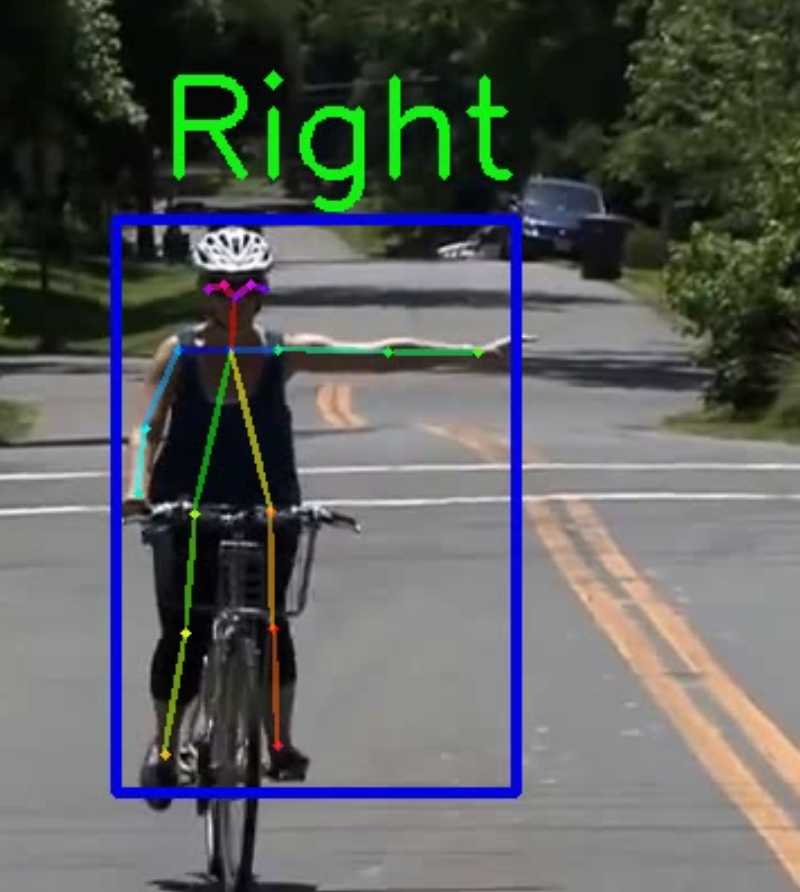}
\caption{Examples of correct predictions in YouTube images (cropped from the original images).}
\label{fig:YTOK}
\end{figure*}

\begin{figure*}
\centering
  \includegraphics[width=0.152\textwidth]{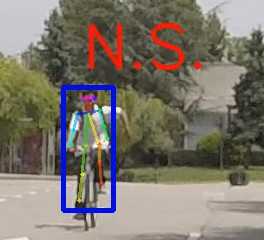}
  \includegraphics[width=0.150\textwidth]{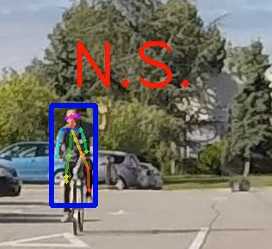}
  \includegraphics[width=0.166\textwidth]{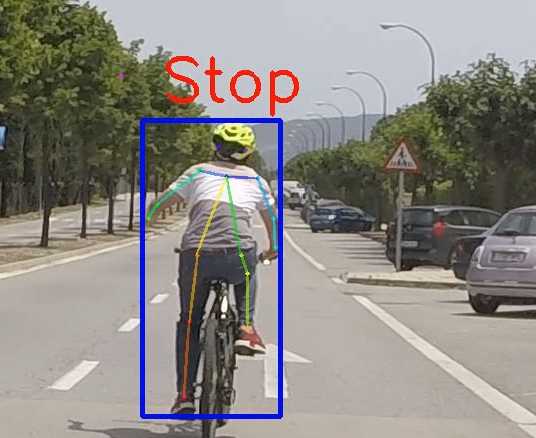}
  \includegraphics[width=0.147\textwidth]{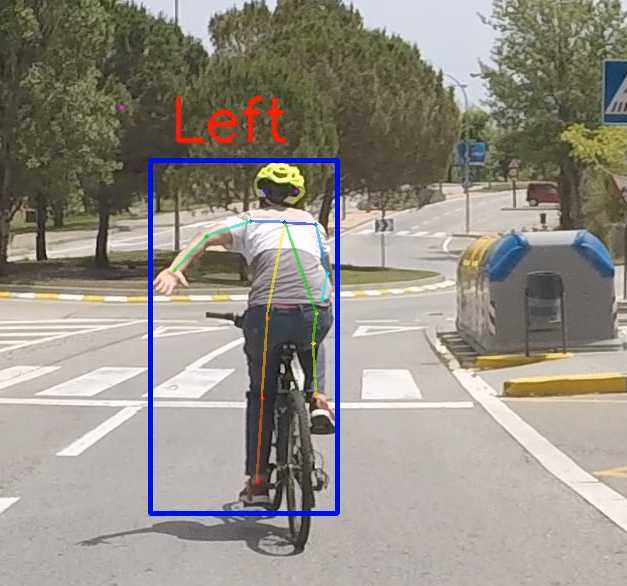}
  \includegraphics[width=0.156\textwidth]{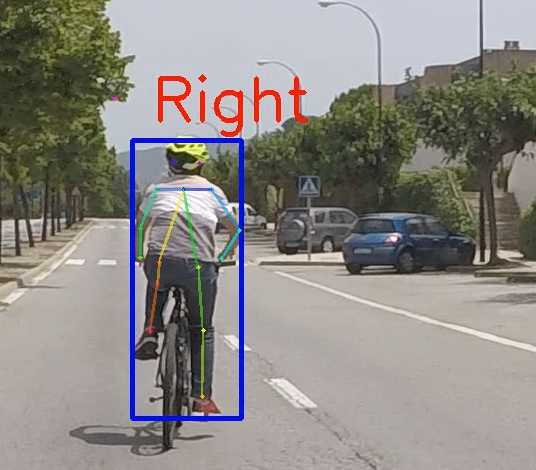}
\caption{Wrong predictions in CASR for cyclist indications (cropped from the original images). 'N.S.' stands for \emph{no sign}.}
\label{fig:wrongCASR}
\end{figure*}

\begin{figure*}
\centering
  \includegraphics[width=0.140\textwidth]{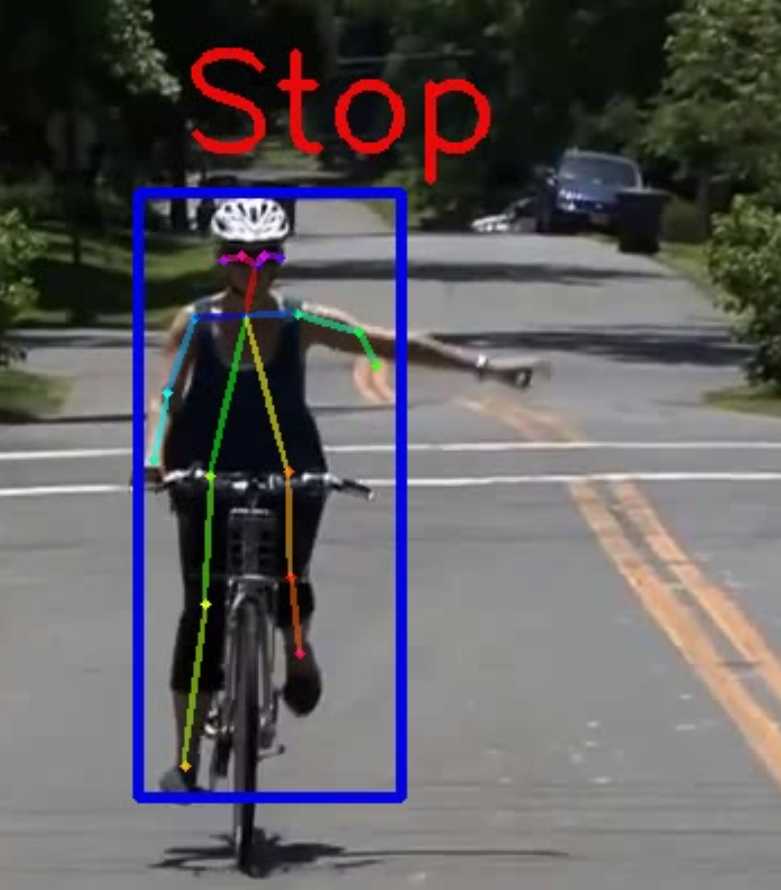}
  \includegraphics[width=0.153\textwidth]{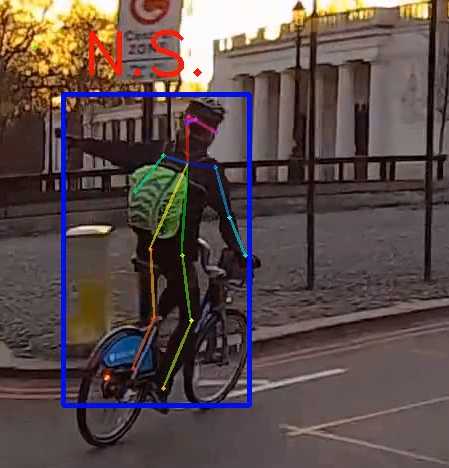}
  \includegraphics[width=0.167\textwidth]{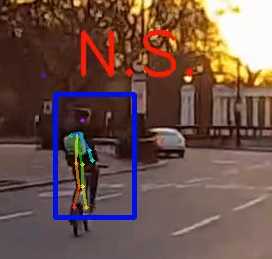}
  \includegraphics[width=0.139\textwidth]{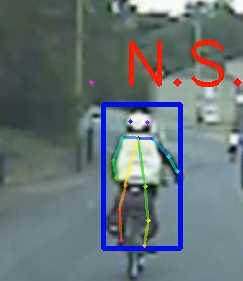}
  \includegraphics[width=0.170\textwidth]{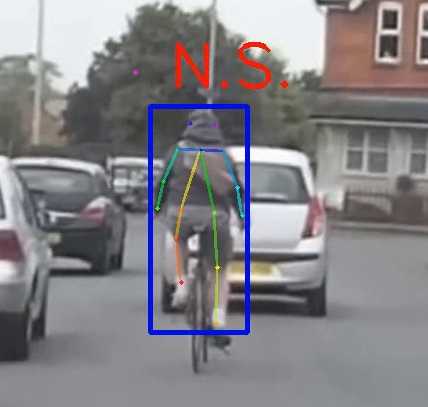}
\caption{Wrong predictions in YouTube images (cropped from the original images). 'N.S.' stands for \emph{no sign}.}
\label{fig:wrongYT}
\end{figure*}

\subsection{Results} 
Table \ref{tab:accuracy_T1T14less} shows the quantitative results for $T=1$ (single frame), and $T=14$ (roughly half a second), respectively. These confirm the effectiveness of the proposed method with relatively high accuracy values, which are quite stable (very low standard deviation). Testing in the YouTube videos is more challenging, but still the accuracy is remarkable since we trained the model on CASR cyclists. In both cases, aggregating temporal information does not help significantly to boost performance; which can be expected since it is already possible to understand what the intentions of the cyclists are looking at a single frame. Still, analyzing more frames can help to stabilize the classification output as we are going to see. In order to confirm this, we force errors in the keypoints as we did for pedestrian intention recognition (see Table \ref{tab:accuracy}).
Table \ref{tab:keypointNoiseResults} compares overall accuracies for noises of $20\%$ and  $30\%$, both for T=1 and T=14. As we can see, only $30\%$ causes an appreciable drop on performance for T=1,  which is avoided up to a large extent thanks to the multiframe setting, {\ie} T=14.   

Figures \ref{fig:leftOK}-\ref{fig:YTOK} present examples of correct results for T=14. The blue BB is the detection. Note how the predictions works for forward and backward faced pedestrians, even if they carry a bag in the back, and at different distances (bigger characters correspond to further away detections). \Fig{wrongCASR} shows some isolated frames with wrong predictions for CASR and T=14. From left to right, the two first cases correspond to oncoming cyclists indicating the intention of stopping and turning to their right (left in vehicle-centric coordinates), but no action is recognized because the detection BBs left the arms out affecting the fitting of the skeleton. This could be solved by systematically augmenting the BB size which is taken as area of interest by the skeleton fitting procedure, at testing time. In the third case the system confuses a future turn left with a stop indication, however, this is the case only at the starting of the action because it is not really clear what the cyclist is going to indicate. The next frames make it clear so that the system actually predicts the proper maneuver. In the fourth case, the system recognises that the cyclist is indicating an action, however, a stop sign is confused with a turn left, which happens because of the relatively straight position of the arm. In this case, we are able to understand the stop indication because of the hand, which is not involved in the analysis of the image. Therefore, this may suggest that a via to explore in future work could be to analyse the area of the image in the extreme of the fitted arms. In the last case, we cannot see any action in this particular frame, while the system indicates a right turn. In fact, in the previous frames, the cyclist actually indicates a right turn; thus, overall this error is more due to the fact that annotating the starting and ending of a given action can have a couple of frames of ambiguity. Therefore, in practice, not detecting any action in this frame or a right turn probably must be considered as correct. \Fig{wrongYT} shows error cases in the YouTube videos for T=14. From left to right, one case due to having the cyclist arm indicating the action out of the BB, two cases due to a bad fitting of the skeleton because adverse conditions (bag in the back, narrow BBs and low contrast arm-background), and two cases where the action has just started and it is not yet clear enough ({\eg} the last case is just the starting of the left-turn action correctly classified in the left example of \Fig{YTOK}). 
We also checked the confusion matrices for CASR results, T=1 and T=14. We did not find any particular confusion pattern between classes, neither for noise free experiments, nor with 20\%/30\%  noise.

Finally, we assess the most important features for the RF classifier. Table \ref{tab:relevanceCyclistFeaturesT1} shows the case T=1, for its \emph{Best} classifier in Table \ref{tab:accuracy_T1T14less}. Most features correspond to angles defined by either a keypoint from neck/shoulders/waist/legs and two keipoints from arms ({\eg} $\Theta(4,6,7)$), or two keypoints from the former set and one from the later ({\eg} $\Theta(3,2,7)$). Distances between these two sets are also among the most relevant ({\eg} $\mbox{L}(8,7)$). Table \ref{tab:relevanceCyclistFeaturesT14} is for T=14 and its \emph{Best} classifier of Table \ref{tab:accuracy_T1T14less}. We see how current frame ({\ie} frame 14 of the temporal sliding window) mainly contributes with angle-based futures, which is coherent with the results of T=1; {\ie} to favour early recognition despite using more frames. We see also how there are many distance features between neck/shoulder/waist/leg and arm keipoints ({\eg} $\mbox{L}^8(8,7),\mbox{L}^9_x(10,1)$), most of them are concentrated in the middle of the temporal window (from frame 5 to 10 we find 12 feature-based features from the top-25, from frame 12 to 14 we find 6) which makes classification more stable once the cyclist has indicated the sign for $\sim200-300\mbox{ms}$. Note also how for cyclists, 25 features are the $\sim2\%$ for T=1, and $\sim0.15 \%$ for T=14.

\begin{table}
\caption{For T=1, top-25 most relevant cyclist skeleton-based features from left-to-right and top-to-bottom.}
\vspace{-0.3cm}
\centering
\begin{scriptsize}
\[
\begin{array}{ccccc}
\toprule
\Theta(4,6,7)  & \Theta(9,6,7)     & \mbox{L}(8,7)  & \mbox{L}(10,7) & \mbox{L}_x(10,1) \\
\Theta(3,6,7)  & \mbox{L}_x(10,11) & \mbox{L}(10,6) & \Theta(3,2,7)  & \Theta(6,3,7)    \\
\mbox{L}(3,2)  & \Theta(11,5,6)    & \Theta(4,7,6)  & \Theta(10,6,7) & \Theta(5,6,13)   \\
\Theta(5,6,12) & \Theta(8,6,7)     & \Theta(11,6,7) & \Theta(3,7,6)  & \Theta(4,2,7)    \\
\Theta(6,2,7)  & \Theta(12,6,7)    & \Theta(5,6,10) & \mbox{L}(5,9)  & \Theta(3,6,12)   \\
\end{array}
\]
\end{scriptsize}
\label{tab:relevanceCyclistFeaturesT1}
\end{table}

\begin{table}
\caption{Analogous to Table \ref{tab:relevanceCyclistFeaturesT1} for T=14.}
\vspace{-0.3cm}
\centering
\begin{scriptsize}
\[
\begin{array}{ccccc}
\toprule
\mbox{L}^{14}(8,7)  & \mbox{L}^{12}_x(10,1) & \mbox{L}^8(8,7)       & \mbox{L}^9_x(10,1)  & \Theta^{14}(12,5,6) \\
\mbox{L}^5_x(10,1)  & \mbox{L}^{10}_x(10,1) & \mbox{L}^6_x(10,1)    & \mbox{L}^{13}(5,9)  & \mbox{L}^6(8,7)     \\
\mbox{L}^{13}(10,6) & \mbox{L}^2_x(10,1)    & \mbox{L}^{13}_x(10,1) & \Theta^{14}(3,6,2)  & \Theta^{12}(2,3,7)  \\
\mbox{L}^8_x(10,1)  & \Theta^{14}(8,6,7)    & \Theta^{12}(4,2,7)    & \mbox{L}^7_x(10,1)  & \mbox{L}^{12}(5,9)  \\
\mbox{L}^8(5,9)     & \Theta^{14}(11,5,6)   & \mbox{L}^7(4,11)      & \mbox{L}^9_x(10,11) & \mbox{L}^{10}(8,11) \\
\end{array}
\]
\end{scriptsize}
\label{tab:relevanceCyclistFeaturesT14}
\end{table}

\section{CONCLUSION}
\label{sec:conclusions}
We have evaluated a monocular vision-based pipeline for recognition of VRU intentions. We have addressed the pedestrian crossing/not-crossing problem in naturalistic driving conditions (JAAD dataset). We have addressed the recognition of cyclist arm signs, in this case elaborating our own dataset, CASR. In both cases, we have applied the same procedure since our work hypothesis is that human skeletons fitted on 2D images already convey core  information for VRU intention recognition. The obtained results support this hypothesis, since by analysing features of the the fitted skeletons over a relatively small temporal sliding window ($\sim500\mbox{ms}$), the recognition task is performed with a high accuracy. We have showed quantitative results supporting this claim, and we have also brought qualitative results (correct recognition cases, current failure cases, top-features driving recognition) illustrating the reasons. Other researchers can use our approach as part of a modular perception pipeline, as affordances on end-to-end driving models \cite{Sauer:2018}, or as additional cue on systems performing 3D trajectory prediction \cite{Kooij:2019}.

\ifCLASSOPTIONcaptionsoff
  \newpage
\fi



%

\bibliographystyle{./IEEE-ITS/IEEEtran}

\bibliography{IEEEabrv,root}

%
\vspace{-1.4cm}
\begin{IEEEbiography}[{\includegraphics[clip=true, trim = 0 10 0 30, height=1.17in, keepaspectratio]{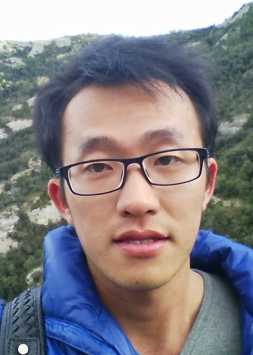}}]{Zhijie Fang}
received his B.S. degree in Automation from South China Agricultural University, Guangzhou, China, in 2011. He received his M.S. degree in Control Science and Control Engineering from South China University of Technology, Guangzhou, China, in 2014. And he is currently a Ph.D. Candidate in the Department of Computer Science at the Universitat Aut\`{o}noma de Barcelona (UAB). His research interests are in the areas of computer vision and pedestrian behavior analyses.
\end{IEEEbiography}
\vspace{-1.6cm}
\begin{IEEEbiography}[{\includegraphics[clip=true, trim = 0 45 0 5, height=1.17in, keepaspectratio]{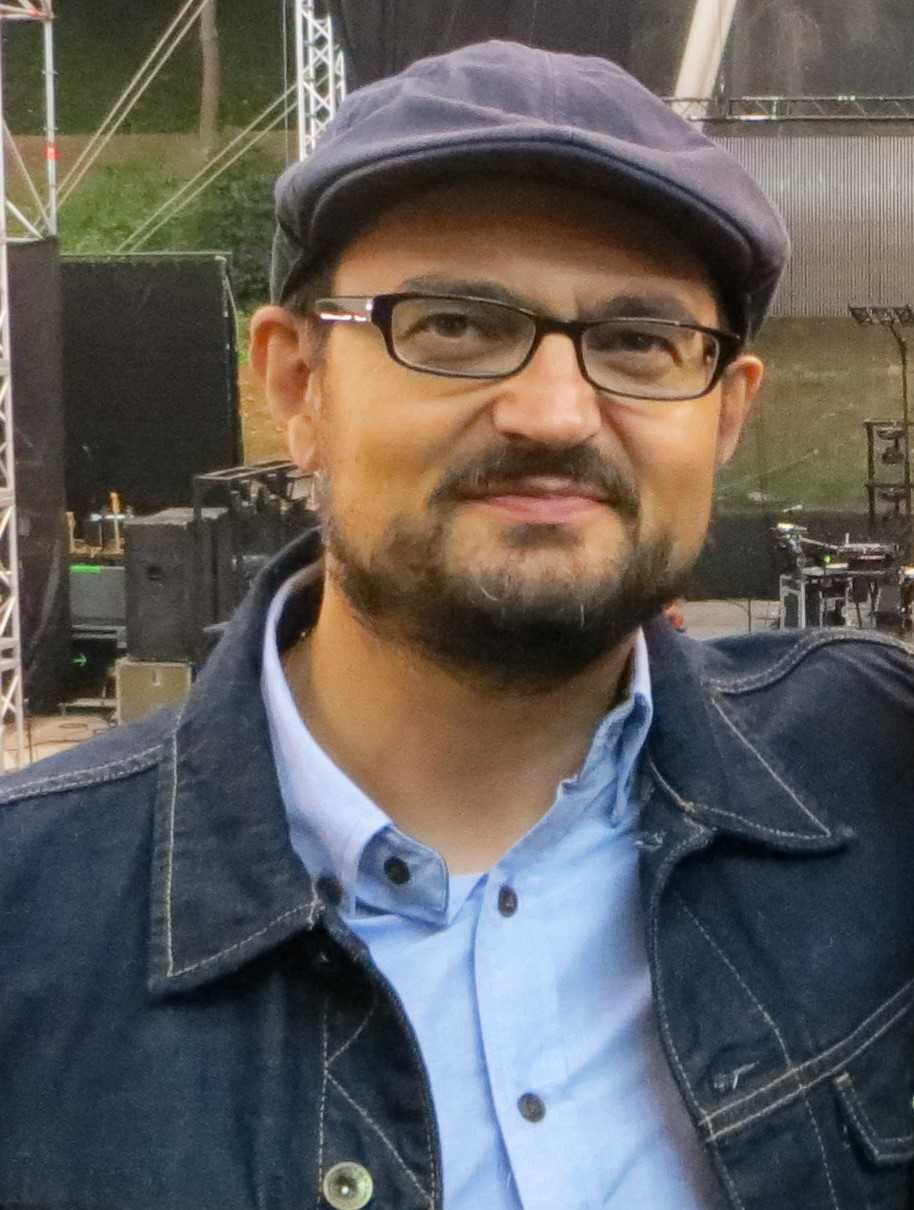}}]{Antonio M. L\'opez} is the PI of the Autonomous Driving lab of the Computer Vision Center (CVC) at the UAB, where he has also a tenure position as associated professor of the Comp. Sci. Dpt. Antonio has a long research trajectory  at the intersection of computer vision, computer graphics, machine learning and autonomous driving. Antonio has been deeply involved in the creation of the SYNTHIA dataset and the CARLA open-source simulator. He is actively working with industry partners to bring state-of-the-art techniques to the field of autonomous driving. Antonio is granted by the Catalan ICREA Academia program.
\end{IEEEbiography}







\end{document}